\documentclass[10pt,letterpaper,compsoc,conference]{iiswc26}

\usepackage{cite}
\usepackage{amsmath,amssymb,amsfonts}
\usepackage{graphicx}
\usepackage[dvipsnames]{xcolor}
\usepackage[final]{microtype}
\usepackage[italic]{mathastext}
\usepackage{libertine}
\usepackage[T1]{fontenc}
\usepackage{textcomp}
\usepackage[varqu,varl]{zi4}
\usepackage[all]{nowidow}
\usepackage[auth-lg,affil-it]{authblk}
\usepackage[keeplastbox]{flushend}
\usepackage{fancyhdr}
\usepackage{booktabs}
\usepackage{makecell}
\usepackage{multirow}

\usepackage{tikz}
\usepackage{algorithmicx}
\usepackage{algpseudocode}
\usepackage{algorithm}
\usepackage{subcaption}
\usepackage{float}
\usepackage{xspace}
\usepackage{enumitem}
\usepackage{listing}
\usepackage{verbatim}

\newcommand{\owl}{OWL\xspace}
\newcommand{\mirothinker}{MiroThinker\xspace}
\newcommand{\dataset}{GAIATrace\xspace}
\newcommand{\sys}{Vidur-Agent\xspace}
\newcommand{\mainllm}{Main-LLM\xspace}
\newcommand{\subllm}{Sub-LLM\xspace}
\newcommand{\mainllms}{Main-LLMs\xspace}
\newcommand{\subllms}{Sub-LLMs\xspace}
\fancypagestyle{firstpage}{
  \fancyhf{}
  
  \fancyfoot[C]{\thepage}
}

\begin{document}

%% EDIT TITLE BELOW

\title{Characterizing How Complex Agentic AI Systems Handle General Tasks: A Trace-Based Simulation Study}
% \title{Characterizing How Complex Agentic AI Systems Handle General Tasks: A Trace-Based Simulation Study}

%Characterizing How Autonomous Agentic AI Systems Handle General Tasks: A Trace-Based Simulation Methodology

%How Autonomous Agentic AI Systems Handle General Tasks: A Trace-Based Simulation Study

%Characterizing How Autonomous Agentic AI Systems Handle General Tasks: A Trace-Based Simulation Study

%% DO NOT EDIT THE FOLLOWING

%\renewcommand\Authsep{\qquad}
%\renewcommand\Authand{\qquad}
%\renewcommand\Authands{\qquad}

%% EDIT AUTHOR LIST BELOW

% \author{Donghwan Kim}
% \author{Prakhar Singh}
% \author{Younghoon Min}
% \author{Jongryool Kim}
% \author{Jongse Park}
% \author{Kiwan Maeng}
% \affil{Full Name of Awesome School}

% \author{}

%% ALTERNATIVE FORMAT FOR MULTIPLE SCHOOLS:
%% 
\author[1*]{Donghwan Kim}
\author[1]{Prakhar Singh}
\author[2]{Younghoon Min}
\author[2]{Jongryool Kim}
\author[3]{Jongse Park}
\author[1*]{Kiwan Maeng}
% \author[1]{Author4 Name}
\affil[1]{The Pennsylvania State University}
\affil[2]{SK hynix Inc.}
\affil[3]{KAIST}
\affil[*]{\{djk6434,kvm6242\}@psu.edu}

\maketitle

\thispagestyle{empty}   % first page: maketitle already set 'plain'
\pagestyle{empty}       % all following pages

\begin{abstract}
Agentic AI completes tasks through iterative planning, tool use, and reasoning based on observed outcomes. Despite its popularity, its system-level behavior remains poorly understood---particularly for complex datasets and agent architectures---owing to highly non-deterministic execution, prohibitive evaluation costs, and limited visibility into proprietary models.
This paper presents \dataset, the first token-level trace dataset of two state-of-the-art agentic systems (\mirothinker and \owl) running GAIA, a benchmark composed of a heterogeneous mix of general-purpose tasks. Unlike prior trace datasets, \dataset captures full reasoning tokens, task-level structures, and activities of every major participating LLMs, enabling in-depth systems research.
Complementing the dataset, we present \sys, a trace-driven simulator that can replay \dataset to perform reproducible, low-cost system evaluation across diverse simulated environments.
Using both artifacts, we characterize how modern agentic systems handle general tasks and how various system design choices shape their behavior, yielding several unique findings.

\end{abstract}

\section{Introduction}
\label{sec:introduction}

%\kwm{TODO
%\begin{enumerate}
%    \item \textbf{High priority TODOs}
%    \begin{enumerate}
%        \item Add background and related work. Especially, add explanations + figures on how MiroThinker and OWL works.
%        \item Add representative traces.
%        \item Briefly show how much compute is wasted.
%    \end{enumerate}
%\end{enumerate}
%}

\emph{Agentic AI}---autonomous pipelines in which LLMs plan, invoke tools, observe results, and iterate until a task is complete~\cite{yao_2023_react, miromind_2026_mirothinker, hu_2025_owl, fourney_2024_magentic_one, su_2026_miroflow}---is a popular new paradigm with the potential to tackle open-ended tasks without human intervention.
Yet, the community's understanding of its system characteristics remains thin. A recent study~\cite{kim_2026_cost} offered valuable early insights but examined only a narrow slice of the design space (Section~\ref{subsec:bg_agentic_ai}).
More complex setups---multi-agent/model systems tackling general tasks with a broad range of tools---remain largely uncharted, because bringing up such systems and running controlled experiments are expensive and challenging.
%, and their highly non-deterministic execution makes controlled experimentation difficult.

To close this gap, we present \emph{\dataset} and \emph{\sys}, a dataset and simulator that together enable cost-efficient and reproducible experimentation on complex agentic AI systems.
\dataset is a trace dataset capturing the behavior of two state-of-the-art agentic systems---\mirothinker~\cite{miromind_2026_mirothinker} and \owl~\cite{hu_2025_owl}---as they solve a heterogeneous mix of general tasks from the GAIA benchmark~\cite{mialon_2023_gaia}. Unlike existing traces~\cite{qin_2025_mooncake, miroverse}, \dataset preserves task-level structure (ordered queries grouped by task), full visibility into reasoning tokens, and traces from every participating LLM, including auxiliary summarization and extraction models.
\sys is a trace-driven simulator that extends the Vidur~\cite{agrawal_2024_vidur} LLM simulator with support for agentic execution: heterogeneous models, KV and prefix caching~\cite{kwon_2023_efficient, zheng_2024_sglang}, prefill--decode disaggregation~\cite{zhong_2024_distserve, patel_2024_splitwise}, and advanced scheduling policies.
Together, they allow agentic AI system simulation to a degree that was impossible before.
%with existing datasets and simulators.

Our characterization studies based on these yield several new findings, organized as takeaways throughout the paper. We highlight a few below (not a comprehensive list):
\begin{itemize}[topsep=2pt, leftmargin=*, itemsep=0pt]
\item \textbf{Agent behaviors are highly heterogeneous.} \dataset reveals far more diverse patterns than what prior work~\cite{kim_2026_cost} observed, driven by heterogeneity in workloads, agents, models, and tools. Its traces deviate substantially from the simple monotonically-increasing prefill pattern observed in simpler setups (\emph{e.g.}, ReAct~\cite{ yao_2023_react}).

\item \textbf{Popular SLA metrics are often misaligned with agentic AI.}
Certain designs significantly worsen per-query metrics such as tail time-to-first-token (TTFT) or time-per-output-token (TPOT), yet improve per-task latency.
Unlike chatbot LLMs---where each query faces a user and must meet strict TTFT/TPOT requirements---most tokens in agentic AI are never read by a human. This suggests that prior optimizations targeting per-query metrics~\cite{zhong_2024_distserve, agrawal_2024_sarathi_serve, wu_2023_fast} may not be optimal for agentic AI.

\item \textbf{System bottlenecks shift with load.} The dominant bottleneck changes with task arrival rate, making static hardware provisioning brittle. A GPU allocation that is near-optimal at one arrival rate can be substantially worse at another, motivating dynamic reconfiguration.

\item \textbf{Prefix caching is far more impactful than previously reported, but unevenly.}
In our study, end-to-end task latency improved by 1.13--3.21$\times$ with prefix caching, which is substantially larger than prior reports (\emph{e.g.}, 15.7\%~\cite{kim_2026_cost}). Yet the benefit is uneven, and some components even experience a \emph{degraded} TTFT. 
The finding again shows how task latency and per-query metrics can misalign.
%The finding again underscores that end-to-end task latency, not per-query metrics, must be the first-class metric in agentic AI.
\end{itemize}

These findings argue that future agentic AI will exhibit characteristics that diverge from what today's simpler agents and workloads display, motivating deeper study of complex agentic systems (multi-agent, multi-model, and multi-tool) and workloads. We hope that \dataset, \sys, and our initial study will be a useful foundation for future research in this direction.

\section{Background and Motivation}
\label{sec:background}
\subsection{LLM Serving Systems}
\label{sec:bg_llm_inference}

Large language models (LLMs) are transformer architectures~\cite{vaswani2017attention} that generate text autoregressively in two phases. The \emph{prefill} phase processes the input prompt (\emph{i.e.}, \emph{context}) in a single pass and produces the key-value (KV) cache for each token in the input, while the \emph{decode} phase generates output tokens one at a time, with each new token attending to the KV cache of all preceding tokens.
%
%The two phases have very different characteristics: 
Prefill is compute-bound, and a single long input can saturate the GPU, whereas decode is memory-bound and can benefit from batching.
%, so decode phase from several queries are typically batched together to improve hardware utilization.
%
%\emph{Reasoning} is a unique 
\emph{Reasoning} is a special type of decode where LLMs produce an extended chain of intermediate ``thinking'' tokens to improve its answer.
Many LLMs allow reasoning to be toggled on or off.
For safety and proprietary reasons, closed-source models do not reveal the reasoning tokens.
%, exposing only the final answer.
%

Several techniques have emerged to optimize prefill and decode.
\emph{Prefix caching}~\cite{he_2026_efficient, zheng_2024_sglang} avoids redundant prefill computation by reusing the KV cache across queries that share a common prefix (\emph{e.g.}, a system prompt or a conversation history).
When prefill and decode run on the same GPU~\cite{yu_2022_orca, kwon_2023_efficient, agrawal_2023_sarathi, agrawal_2024_sarathi_serve, holmes_2024_deepspeedfastgen}, their distinct compute patterns interfere with each other.
\emph{Prefill-decode (PD) disaggregation}~\cite{hu_2025_shuffleinfer, patel_2024_splitwise, zhong_2024_distserve} solves this problem by running them on separate devices connected by high-bandwidth interconnects.
%for better hardware utilization.
%
When using PD disaggregation, each phase can use a different number of GPUs and a different parallelism strategy, such as tensor parallelism (TP)~\cite{narayanan_2021_megatron} or pipeline parallelism (PP)~\cite{harlap_2018_pipedream, huang_2019_gpipe}.
A common approach is to profile a representative workload offline and statically configure the GPUs for each phase~\cite{zhong_2024_distserve}.
%Section~\ref{sec:related} discusses other important research on LLM serving systems.

%\kwm{I don't know these citations much, so just wrote something super general \& random. Please refine, and may add more citations @Donghwan}
%
Over the years, numerous systems were proposed for efficient LLM serving~\cite{yu_2022_orca, kwon_2023_efficient, zheng_2024_sglang, agrawal_2023_sarathi, agrawal_2024_sarathi_serve, zhong_2024_distserve, patel_2024_splitwise, qin_2025_mooncake, pan_2025_instattention, liu_2025_lmcache, sheng_2023_flexgen, she_2026_laps, anhyankar_2024_infercept, yoon_2025_tract, goel_2026_qoserve, hidayetoglu_2026_shift_parall, lin_2026_bullet, zhao_2026_blendserve, yi_2026_pat, du_2026_bitdecoding, zhang_2025_jenga, yu_2025_prism, xia_2025_skywalker} and agentic AI serving~\cite{lin_2024_parrot, chaudhry_2025_murakkab, biswas_2026_sutradhara, Tan_2025_teola, raghavan_2025_alto, zhang_2025_jitserve, li_2026_continuum, qin_2025_mooncake, kang_2026_thunderagents, pan_2025_kvflow, luo2025autellix, lin_2024_parrot, bian2025tokencake, sheng_2024_fairness, chen_2026_concur, he_2026_efficient, woo_2026_icarus}.
These efforts are largely orthogonal to this paper.
% \kwm{I don't know these work, so just wrote something random. Please refine, and may add more citations @Donghwan}
%introducing optimizations for efficiently maintaining KV cache~\cite{icarus, li_2026_continuum, qin_2025_mooncake, kang_2026_thunderagents, pan_2025_kvflow} or more advanced scheduling policies~\cite{luo2025autellix, lin_2024_parrot, bian2025tokencake, sheng_2024_fairness}. 

\subsection{(Autonomous) Agentic AI}
\label{subsec:bg_agentic_ai}

%Agentic AI refers to a new paradigm in which LLMs operate not just as text generators, but as decision-makers that plan, invoke external tools, observe results, and iterate toward a goal.
%
Agentic AI is an overloaded terminology in the community without a clear definition.
Following the definition from Anthropic~\cite{anthropic_blog_agents}, we focus on \emph{autonomous} systems that rely on the LLM to dynamically decide how to solve the given problem---generating plans, calling tools and interpreting their results, invoking other LLMs, and judging when the task is complete---without a human in the loop.
We do not consider systems whose control flow is statically designed by human (which Anthropic refers to as ``workflows'' or ``augmented LLMs''~\cite{anthropic_blog_agents}).

%
%\%emph{Autonomous} agentic AI is the most sophisticated form in this paradigm~\cite{todo}.
%
%Unlike simpler agentic systems (which, some refer to as ``workflows'' or ``augmented LLM''~\cite{anthropic_blog} to distinguish from autonomous agentic AI) whose control flow is statically designed by a human~\cite{todo}, autonomous agentic AI relies on the LLM itself to dynamically decide how to solve the given problem---generating plans, calling tools and interpreting their results, invoking other LLMs, and judging when the task is complete---without a human in the loop.
%The term ``agent'' in the literature is often overloaded and not clearly defined.
%
%In this paper, we call each LLM that serves a particular role (\emph{e.g.}, planning, web browsing, or coding) is called an \emph{agent}, and a single LLM instance can serve as multiple agents.
%
%This paper focuses on autonomous agentic AI; throughout, ``agentic AI'' refer to autonomous agentic AI.

\paragraph{{Design of Agentic AI}}
How to architect an agentic AI system remains an open question under active research.
\emph{Early agentic systems}~\cite{yao_2023_react, zhou_2023_lat, kim2023llmcompiler, miromind_2026_mirothinker, li_2025_search_o1, wu_2025_webdancer} adopted simple structures in which an LLM repeatedly alternates between planning, tool invocation, and replanning based on tool outcomes.
At each turn, the history of prior steps (plans and tool results) is appended to the context, causing the input length to grow monotonically.
We refer to such systems as \emph{ReAct-like}, after the canonical example, ReAct~\cite{yao_2023_react}.
%
%ReAct-like systems are amenable to prefix caching, since the KV cache for most of the input tokens (past history) has already been computed in previous turns.
%
%Subsequent designs preserved this high-level structure while addressing its limitations; MiroThinker~\cite{miromind_2026_mirothinker}, for instance, retains only the most recent $k$ histories to bound context growth.
%
\emph{More recent agentic systems}~\cite{su_2026_miroflow, hu_2025_owl, fourney_2024_magentic_one, hu_2026_flowsearch, yang_2025_agentnet, xie_2024_aimetropolis, li_2023_camel, hong_2024_metagpt, asgar_2025_efficient, wu_2023_autogen} explore richer architectures in which multiple agents are organized hierarchically and communicate to solve problems collaboratively.
The same LLM can serve as multiple conceptually different agents (\emph{i.e.}, role-play~\cite{li_2023_camel}).

\begin{figure}[t]
\centering
\includegraphics[width=0.85\columnwidth]{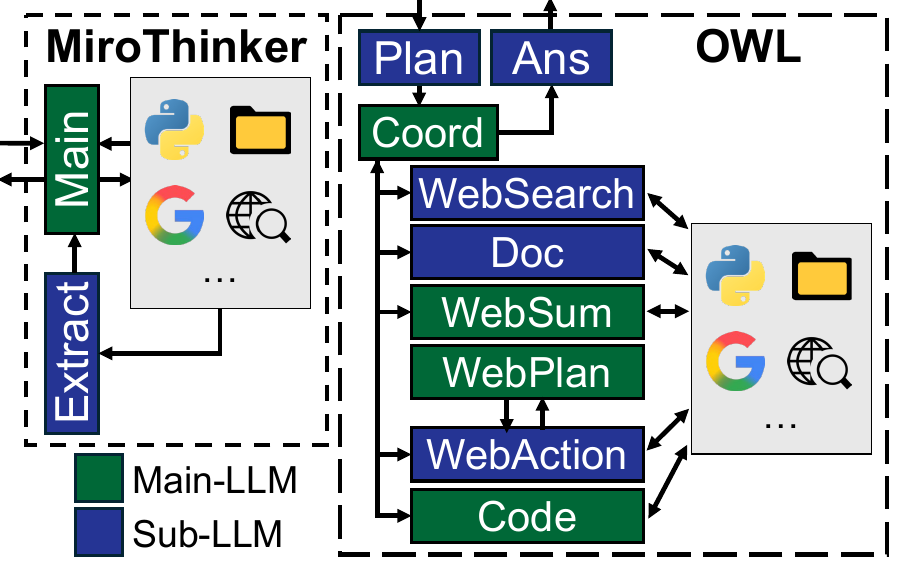}
\caption{Two agentic AI systems, \mirothinker~\cite{miromind_2026_mirothinker} and \owl~\cite{hu_2025_owl}.
Some details omitted for simplicity.
}
\label{fig:overview}
\end{figure}

Figure~\ref{fig:overview} illustrates two representative systems from each family that we will study,  \mirothinker~\cite{miromind_2026_mirothinker} and \owl~\cite{hu_2025_owl}.
%\mirothinker~\cite{miromind_2026_mirothinker}, which is based on a simplistic, ReAct-like design, and \owl~\cite{hu_2025_owl}, which has a more complex structure.
%
\emph{\mirothinker}~\cite{miromind_2026_mirothinker} (left) is a simple, ReAct-like system. The main LLM (``Main'' in Figure~\ref{fig:overview}) generates plans, invokes tools, observes the result, and refine its plans further.
It occasionally gets help from another LLM (``Extract'' in Figure~\ref{fig:overview}) that summarizes and extracts important data from a large, web-scraped text.
While simple, \mirothinker is still more complicated than many ReAct-like systems prior works studied, as it involves multiple LLMs (``Main'' and ``Extract'') and tools (our setup used 4 tools; Table~\ref{tab:tool_stats}).
\emph{\owl}~\cite{hu_2025_owl} (right) has a more complex structure. There exist multiple agents that are specialized in a certain task working under the master coordinator, each generating their own sub-plan, maintaining memory, and making decisions.
As different colors indicate, different LLM architectures are used for different agents.
We omit the details of each system, which can be found in the respective papers~\cite{miromind_2026_mirothinker, hu_2025_owl}.

\paragraph{{Agentic AI Benchmarks}}
Agent AI benchmarks span a wide range of domains.
Many popular benchmarks consist of homogeneous tasks, such as question answering~\cite{yang2018hotpotqa}, function call selection~\cite{patil_2025_bfcl}, web browsing~\cite{zhou_2024_webarena}, code generation~\cite{chen_2021_codex, jimenez_2024_swebench}, and scientific reasoning~\cite{hendrycks_2021_math, rein_2024_gpqa}.
These benchmarks can typically be solved with only one or two tools~\cite{kim_2026_cost}, and each workload exhibits a relatively homogeneous computational pattern.
Other benchmarks~\cite{mialon_2023_gaia, yoran_2024_assistantbench, qi2025agentif} instead comprise mixed real-world tasks that require a more diverse set of tools and exhibit more heterogeneous compute patterns.
This paper focuses on one such general benchmark, GAIA~\cite{mialon_2023_gaia}.
%
%Agentic AI benchmarks consist of \emph{tasks}, which are higher-level request from users.
%Completing a task involves multiple LLM \emph{queries} (pair of prefill and decode phases) to be served, each performing a sub-task to complete the higher-level task.

\paragraph{\textbf{Goal: Characterizing the Broader Landscape}}
A recent characterization study~\cite{kim_2026_cost} offered valuable early insights into the system overheads of agentic AI, but it only examined a narrow slice of this vast and rapidly evolving space.
In particular, the prior work~\cite{kim_2026_cost} only studied simple ReAct-like systems on single-domain benchmarks (\emph{e.g.}, MATH~\cite{hendrycks_2021_math} and HotPotQA~\cite{yang2018hotpotqa}) with only one or two tools enabled per benchmark.
Such setups fail to capture the complexities of modern agentic AI: heterogeneous tasks executing concurrently on complex agentic structures with a wide variety of tools.
This paper aims to close the gap by studying a hetorogeneous mixture of workloads running on complex agentic systems (multi-agent, multi-model, and multi-tool), and simultaneously develop a benchmark and simulator for future research in this direction.

\subsection{LLM Simulators}
\label{subsec:bg_simulator}

LLM serving has become a major system optimization problem, motivating the development of simulators~\cite{Cho_2024_llmservingsim, Cho_2026_llmservingsim, agrawal_2024_vidur, zhang_2024_LLMCompass, kim_2025_ador, lin_2025_apex, wu_2025_tokensim, bambhaniya_2026_mist} that evaluate LLM serving efficiency without the cost of full deployment.
%
%Simulators offer flexibility and reproducibility, enabling rapid exploration of design decision across diverse hardware and software configurations.
Vidur~\cite{agrawal_2024_vidur} is an LLM simulator built by Microsoft Research that predicts the serving efficiency under various models, GPU hardware, parallelization strategies, batching, and scheduling policies, through a learned decision tree.
The accuracy of Vidur has been verified, and several system design studies have been built on top of Vidur~\cite{agrawal_2024_sarathi_serve, agrawal_2026_revati, goel_2026_qoserve}.
However, Vidur operates at the individual query level and does not model the multi-turn structure and inter-turn dependencies of agentic AI. We extended Vidur to build \sys (Section~\ref{sec:simulator_design}).

Several other LLM simulators have also been proposed, including LLMServingSim and 2.0~\cite{Cho_2024_llmservingsim, Cho_2026_llmservingsim}, TokenSim~\cite{wu_2025_tokensim} and APEX~\cite{lin_2025_apex}.
%Our \sys is based on Vidur~\cite{agrawal_2024_vidur}, but our contribution is not tied to it, because \dataset will be compatible with other simulators as well with proper modification.
%
At a different abstraction level, LLMCompass~\cite{zhang_2024_LLMCompass} and ADOR~\cite{kim_2025_ador} focus on simulating architectural optimizations and hardware configurations.
%
%These are general LLM simulators and do not focus on agentic AI.
Our \dataset and the general methodology can be made compatible with these other simulators with proper modification.

\begin{table*}[t]
\centering
\caption{Token-level trace statistics across agentic systems and models. \label{tab:trace_stats}}
\resizebox{\textwidth}{!}{
\begin{tabular}{ll c cc c cc}
\toprule
& & \textbf{Mooncake}~\cite{qin_2025_mooncake}
  & \multicolumn{2}{c}{\textbf{\mirothinker (GAIA-103-Text)}}
  & 
  & \multicolumn{2}{c}{\textbf{\owl (GAIA-165-Val)}} \\
\cmidrule(lr){4-5}\cmidrule(lr){7-8}
& & 
  & \makecell{\mainllm\\(\textit{MiroThinker-1.7-mini})}& \makecell{\subllm\\(\textit{gpt-4o-mini})}
  &
  & \makecell{\mainllm\\(\textit{gpt-oss-120b}} & \makecell{\subllm\\(\textit{gpt-4o})}\\
\midrule
$n_\text{task}$ / Queries
  & & $-$ / 23608
    & 103 / 1491 & 103 / 591
    &
    & 165 / 2669 & 165 / 2737 \\
\midrule
Queries per Task
  & {[min, max]}   & $-$
    & [2, 48]  & [1, 25]
    &
    & [2, 113] & [3, 54] \\
  & mean $\pm$ std & $-$
    & 14.5 $\pm$ 10.3 & 7.3 $\pm$ 5.1
    &
    & 16.2 $\pm$ 12.7 & 16.6 $\pm$ 9.9 \\
\midrule
Prefill (tokens)
  & {[min, max]}   & [891, 126195]
    & [2332, 22727] & [165, 93393]
    &
    & [204, 62813] & [713, 122711] \\
  & mean $\pm$ std & 8596 $\pm$ 1101
    & 11201 $\pm$ 5437 & {17248} $\pm$ 20923
    &
    & 4804 $\pm$ 5626 & {7848} $\pm$ 12908 \\
\midrule
Decode (tokens)
  & {[min, max]}         & [1, 2000]
    & [30, 8419] & [5, 8192]
    &
    & [5, 3790] & [0, 7546] \\
  & mean $\pm$ std (total)  & 182 $\pm$ 242
  326.7±482.8
    & {348} $\pm$ 572 & 190 $\pm$ 530
    &
    & {324} $\pm$ 485 & 93 $\pm$ 173 \\
  & mean $\pm$ std (reason) & $-$
    & 245 $\pm$ 556 & $-$
    &
    & 149 $\pm$ 170 & $-$ \\
\midrule
KV Cache Hit Rate (\%) & intra-task / inter-task
  & $-$ / 59
  & 79.0 / 80.4 & {0.3 / 0.3}
  &
  & 58.0 / 58.7 & 60.4 / 61.3 \\
Reasoning Token Ratio (\%) &
  & $-$
  & 70.3 & $-$
  &
  & 47.0 & $-$ \\
\bottomrule
\end{tabular}
}
\end{table*}

\section{\dataset and \sys: Trace-based Simulation for Agentic AI System Study}

Our goal is to study how complex agentic AI systems like \mirothinker~\cite{miromind_2026_mirothinker} and \owl~\cite{hu_2025_owl} behave when running mixed real-world tasks like GAIA~\cite{mialon_2023_gaia}.
First, we discuss the new trace dataset and trace-based simulator that we built.
In Sections~\ref{sec:trace_characterization} and \ref{sec:simulation}, we discuss the charaterization results using these dataset and simulator.

\subsection{Motivation: Why Trace-based Simulation?}

Studying the behavior of agentic AI systems poses two main challenges.
First, their behavior is highly non-deterministic. 
%While traditional LLMs already exhibit stochasticity, the variability in agentic systems is far more severe. 
For the same task, an agent may issue a web search in one trial---incurring tool call and a heavy prefill overheads---while in another trial, forgo the search and proceed with deep reasoning, incurring high decode overheads instead. This makes the system behavior fluctuate dramatically across trials. Fixing randomness (\textit{e.g.}, via \texttt{random.seed(0)}) does not eliminate this stochasticity, because external tools (\textit{e.g.}, Google search) still introduce non-determinism.
Second, evaluation costs are extremely high, with each task taking minutes to run and uses LLMs with tens to hundreds of billions of parameters. 
%Because completing a task traverses multiple task-specific stages (prefill-heavy ingestion, decode-heavy reasoning, tool invocations, \textit{etc.}), each task must be run end-to-end to correctly capture the system behavior. 
%The high evaluation cost significantly raises the barrier to entry for agentic AI systems research.

To address these challenges and lower the cost of agentic AI systems research, we built: (1) \textit{\dataset}, a trace dataset containing detailed behavioral traces of two representative agentic AI systems (\mirothinker~\cite{miromind_2026_mirothinker} and \owl~\cite{hu_2025_owl}) tackling the GAIA~\cite{mialon_2023_gaia} benchmark, and \textit{\sys}, a trace-driven simulator (built on top of the Vidur~\cite{agrawal_2024_vidur} LLM simulator) that can replay the traces and simulate GPU and tool overheads.
%to quickly estimate end-to-end system behavior. 
\dataset and \sys allow controlled experiments on various simulated hardware setups, enabling reproducible and rapid experimentation at low cost. 
%Building \dataset and \sys required careful modification and adoption of existing frameworks (\owl~\cite{hu_2025_owl}, \miroflow~\cite{su_2026_miroflow}, and Vidur~\cite{vidur}).
We explain in Section~\ref{sec:trace_collection} and Section~\ref{sec:simulator_design} how we developed \dataset and \sys.

\subsection{\dataset: Trace Collection}
\label{sec:trace_collection}

While there are existing traces for agentic AI, none of them entirely fits our need. For example, Mooncake's Tool\&Agent traces~\cite{qin_2025_mooncake} only contain individual {query} information (\emph{i.e.}, pair of prefill and decode tokens) and not higher {task}-level information (completing each \emph{task} involves running multiple LLM \emph{queries} in a specific order).
MiroVerse traces~\cite{miroverse} do not contain reasoning tokens and only contain traces from the main LLM---\emph{e.g.}, trace of models annotated as ``Extract'' on Figure~\ref{fig:overview} is omitted.
For a comprehensive study of all aspects, we built our own trace dataset: \dataset includes \textit{reasoning tokens}, traces from \textit{all the major LLMs} that participate, and contains higher \textit{task-level information}.
%
%We chose two representative autonomous agentic AI systems, \owl~\cite{hu_2025_owl} and \mirothinker~\cite{su_2026_miroflow}, and extracted their behavioral traces to construct \dataset.
%
\dataset is valuable in several aspects:
\begin{enumerate}[topsep=5pt, itemsep=0pt, leftmargin=*]
    \item They offer insights into how SOTA agentic AI systems behave (Section~\ref{sec:trace_characterization}).
    \item They can be replayed in \sys to enable rapid experimentation and prototyping (Section~\ref{sec:simulation}).
    \item They can also be replayed directly on real systems (\textit{e.g.}, frameworks such as vLLM support benchmarking custom trace~\cite{vllm_bench} with little engineering effort, and the LLM portion of our traces can be replayed on these systems) to more accurately estimate runtime overheads.
\end{enumerate}

We collected \dataset on a carefully-engineered setup which balances transparency and task completion accuracy.
%designed through extensive trial-and-error.
%
First, we used open-source LLMs when heavy reasoning was needed to allow full inspection of the reasoning tokens. We refer to such models as \textit{\mainllm} throughout the text (Figure~\ref{fig:overview}, green).
We used gpt-oss-120b for \owl and MiroThinker-1.7-mini for \mirothinker.
%Using inspectable open-source models for reasoning agents is essential, as their reasoning behavior is central to the system's overall characteristics.
%
For non-reasoning LLMs, we used high-quality proprietary models (gpt-4o for \owl and gpt-4o-mini for \mirothinker) to ensure high accuracy. Doing so is acceptable because their behavior is determined entirely by their input and output tokens, both of which are visible to the user. We refer to such models as \emph{\subllm} (Figure~\ref{fig:overview}, blue).
For video- and audio-processing, we used Gemini-2.5-mini and Whisper-2, both of which are proprietary.
Although their internal behaviors are opaque, these are invoked infrequently and contribute negligibly to the overall characterization. 

We prevented the agentic systems from directly finding GAIA answers on the web by blocking access to URLs that contain GAIA question-answer pairs (\emph{e.g.}, \texttt{huggingface.co/}), and manually inspected the traces to ensure they represent real-world problem solving behaviors.
We ran each task at most four times. We annotated successful and failed trials, so that one can include only successful trials (representing a high-quality system with a high success rate) or all the trials (modeling the affect of failed execution as well) for simulation. 
%\dhk{The trace set utilized in our characterization (Table~\ref{tab:trace_stats}) includes one trace per task, while prioritizing successful traces.}
%
\owl ran all the 165 tasks in GAIA, while \mirothinker could only run the 103 tasks that do not involve image, video, or audio.
The respective success rate was 48.5\% for \owl and 59.2\% for \mirothinker, which were slightly lower than SOTA numbers~\cite{miromind_2026_mirothinker, su_2026_miroflow, hu_2025_owl} but roughly in a similar ballpark.
%
%We could not reproduce the exact SOTA accuracy 
%
We could not achieve the SOTA accuracy because (1) we partially rely on open-source models with lower quality and (2) some high accuracy numbers couldn't be reproduced even when using the default configurations (similar reproducibility issues were reported by others as well~\cite{owl_reproduce1, owl_reproduce2}). 

Our \mirothinker trace has 26.9M input tokens and 0.6M output tokens, and 44.0\%, 44.1\% of them are from failed traces, respectively.
\owl trace has 34.3M input tokens and 1.1M output tokens, 63\% and 62\% of them are from failed traces, due to lower accuracy.
The statistics indicate that many processed tokens are wasteful in current agentic AI systems (did not lead to successful task completion), which highlights the importance of improving the accuracy.

\subsection{\sys: Simulator Design}
\label{sec:simulator_design}

We modified Vidur~\cite{agrawal_2024_vidur} and built \textit{\sys}, which can replay \dataset and study its system behavior on various system configurations.
%
%Vidur~\cite{agrawal_2024_vidur} models the execution time of popular LLMs under various GPU architecture through a learned decision tree.
%Vidur been verified that its estimation is highly accurate, and several system design studies have been built on top of Vidur~\cite{agrawal_2024_sarathi_serve, agrawal_2026_revati, goel_2025_niyama, zhong_2024_distserve}.
%
%However, Vidur only focuses on simulating the execution time of single prefill and decode stages of an LLM. We made several modifications to model the end-to-end agentic AI system behavior.
%
First, we added support for dependent execution and tool call overhead simulation, \textit{i.e.}, execution of a query only starts when all the dependent queries and tool calls finishes. 
% We measured various tool calls' latency and modeled them with a \kwm{Gaussian distribution}.
%\dhk{
We measured various tool calls' latency with repeated runs and used the measurements to model their delays.
%inject them as a tool execution delay. 
%Rather than modeling using gaussian distribution ~\ref{todo} We measured different tool latency for unique request, as we observed the execution time varies by request~\ref{tab:tool_stats}.  
%}.
%
Second, we implemented modeling advanced system setups that are popular in agentic AI, including support for heterogneous models~\cite{hu_2025_owl, su_2026_miroflow}, prefill-decode disaggregation~\cite{zhong_2024_distserve, patel_2024_splitwise, hu_2025_shuffleinfer}, KV cache~\cite{qin_2025_mooncake, vllm_Kv_routing} and prefix cache~\cite{vllm_prefix_aware_routing}, load-aware query routing~\cite{xia_2025_skywalker, he_2026_efficient}, and advanced scheduling policies~\cite{wu_2023_fast, luo2025autellix}.
A concurrent work proposed a simulator with similar capabilities~\cite{Cho_2026_llmservingsim}, and we do not claim these additional features as a strong novelty.
Our main contribution lies in the collection of \dataset and its characterization (Sections~\ref{sec:trace_characterization} and \ref{sec:simulation}), and \sys is one essential component enabling the characterization. With proper modifications, we believe other simulators (\textit{e.g.}, \cite{Cho_2026_llmservingsim}) can also be used in the place of \sys.

%1. Agentic Trace support.
%Requests carries an inter request latency representing tool call duration between turns. The next turn is released only after current turn completes plus this gap, accurately modeling Agentic trajectories.
%We have dependencies between requests in single session, so turns are only proceeded when all predecessor requests are complete, supporting DAG-structured multi-turn workloads.

%2. PD Disaggregation wtih Heterogeneous cluster configuration. Replicas can differ in model and parallelism degree. Separate prefill and decode pools are maintained per model, with load-aware prefill routing. KV transfer stalls are explicitly modeled at handoff, tokens not cached on the decode side are converted to latency via intra- or cross-node bandwidth estimates. These open a new simulation axis for studying heterogeneous multi-model deployments under agentic workloads which we showed in our paper.

%3. Advanced Intra-replica scheduling
%We had a deeper analysis on scheduling policies. Three representative policies like request level FCFS, task level FCFS, SJF with starvations shows different perspective. we offer analysis tools as well. We put prediction based scheduling policies (prediction based KV cache eviction) ~ref{continuum} as future work.

\section{\dataset Characterization Study}
\label{sec:trace_characterization}

We first analyze several aspects of \dataset, showing that it captures unique behaviors of agentic AI systems.
%working on general tasks.

%\subsection{Trace Collection Methodology}
%\label{subsec:trace_method}

%\paragraph{Multi Agent}
%The models used by OWL in its original configuration are GPT-4o, GPT-o3-mini, and Claude Sonnet 3.7. Proprietary model APIs provides no visibility into internal reasoning tokens.\ To obtain complete execution traces, we replaced GPT-o3-mini with GPT-oss-120B, an open-weight reasoning model from the same organization whose architecture is publicly documented and whose full token sequences are accessible.

%We ran OWL end-to-end on the GAIA benchmark using GPT-4o and GPT-oss-120B, logging every prompt and completion at the token level. The agent foundation models are based on either reasoning LLM (GPT-oss-120B) or non-reasoning VLM (GPT-4o).

%For multi-modal tools like \texttt{ask\_questions\_about\_video} (backed by Gemini-2.5-mini) and \texttt{ask\_questions\_about\_audio} (backed by Whisper-2) remains proprietary, we model latency using sampled empirical distributions rather than token-level traces.

%\paragraph{Single Agent}
%We utilized MiroThinker-1.7-mini, which is fine-tuned over
%\texttt{Qwen/Qwen3-30B-A3B-Thinking-2507}. We could acquire all the reasoning token from this model. There is one critical tool named  \texttt{scrape\_and\_extract\_info}, which is sending scraped webpage with question, is backed by gpt-4o-mini. As it is LLM-assisted tool, and is the most significant function, we regard this trace as two model traces as well.

\subsection{Overall Statistics}
\label{subsec:workload_overview}

Table~\ref{tab:trace_stats} summarizes the token-level statistics of \dataset.
In this table, we only include one trace per task when there were multiple trials, preferring successful traces.
As discussed in Section~\ref{sec:trace_collection}, both \owl and \mirothinker primarily rely on two types of LLMs: a \textit{\mainllm} with a reasoning capability and a \textit{\subllm} without.
%
%Through extensive trial-and-error (Section~\ref{sec:trace_collection}), we settled on MiroThinker-1.7-mini (\miroflow) and gpt-oss-120b (\owl) as \mainllm, and gpt-4o-mini (\miroflow) and gpt-4o (\owl) as \subllm.
%
We report token statistics for each model separately.
For comparison, we also include Mooncake's Tool\&Agent trace~\cite{qin_2025_mooncake}.
%, a real-world agentic AI trace collected by Moonshot AI.

\paragraph{Number of Tasks and Queries}
Unlike the Mooncake trace, which only reports statistics for individual LLM queries, \dataset additionally captures the higher-level task structure: the dataset consists of multiple high-level tasks, each handled through a series of LLM queries and tool calls. This makes \dataset better suited for studying task-level behavior. As shown in Table~\ref{tab:trace_stats}, \mirothinker completes 103 tasks via 1{,}491 \mainllm queries, 591 \subllm queries, and a number of tool calls (discussed later in Table~\ref{tab:tool_stats}). \owl completes 165 tasks via 2{,}669 \mainllm queries, 2{,}737\subllm queries, and tool calls. Again, \owl can also handle image, video, and audio inputs, covering more tasks.

On average, \owl issues more queries per task (16.2--16.6) than \mirothinker (7.3--14.5).
This is due to how \owl is designed, where multiple agents frequently communicate with each other. This illustrates how agentic AI system design significantly impacts LLM query volume.
The high standard deviation in queries per task further indicates that GAIA tasks vary substantially in complexity and requires different number of LLM queries to complete.
%Mooncake does not reveal such task-level statistics.

\paragraph{Prefill Statistics}
The \mainllms of both systems process a similar number of prefill tokens (4{,}804--11{,}201) per query as Mooncake (8{,}596). \mirothinker's \mainllm sees slightly longer prefills than \owl's, as its ReAct-like design more aggressively accumulates the history.
The standard deviation is notably higher than Mooncake's (5{,}437--5{,}626 vs. 1{,}101), which we attribute to varying task complexity of the GAIA and the higher complexity in the agent architecture.
We also report the ideal KV cache hit rate (assuming unbounded cache size). The \mainllms of \dataset exhibit slightly higher hit rates than Mooncake, with most hits coming from queries within the same task (intra-task). \mirothinker achieves a higher hit rate due to its ReAct-like design, which yields substantial prefill overlap between adjacent queries.
%(shown further in Figure~\ref{fig:representative}).

The \subllm prefill statistics paint a more interesting picture. Since \subllms handle prefill-heavy tasks such as text summarization and image understanding, their average prefill lengths are higher (7{,}848--17{,}248), and their standard deviations are extreme (12{,}908--20{,}923), reflecting wide variation in the size of input texts and images.
Most notably, \mirothinker's \subllm suffers from an \textit{extremely low KV cache hit rate of 0.3\%}. This is because it is primarily used to summarize scraped web pages, and prefill overlap is essentially nonexistent unless the same page is scraped and summarized twice.
These observations indicate that KV cache hit rates can vary significantly across agent designs.

\paragraph{Decode Statistics}
The decode length of models in \dataset is roughly in the similar ballpark with Mooncake (93--348 vs. 182), but again have much higher variance. We additionally observed that nearly half of the generated tokens of reasoning models (\mainllms) are reasoning tokens (47.0\% for \owl, 70.3\% for \mirothinker).
This emphasizes the importance of accurately modeling the impact of reasoning tokens, which is omitted in many existing trace datasets (\textit{e.g.}, MiroVerse~\cite{miromind_2026_mirothinker}). %\dhk{Omission of reasoning tokens in the middle of decode tokens deters the KV cache reuse as prefix mismatches in subsequent turn query.}
%

%
%This point is further emphasized in the next section through a more detailed analysis to each task.

\subsection{Detailed Trace Analysis}
\label{subsec:per_agent_distribution}

\begin{figure}[t]
\includegraphics[width=\columnwidth]{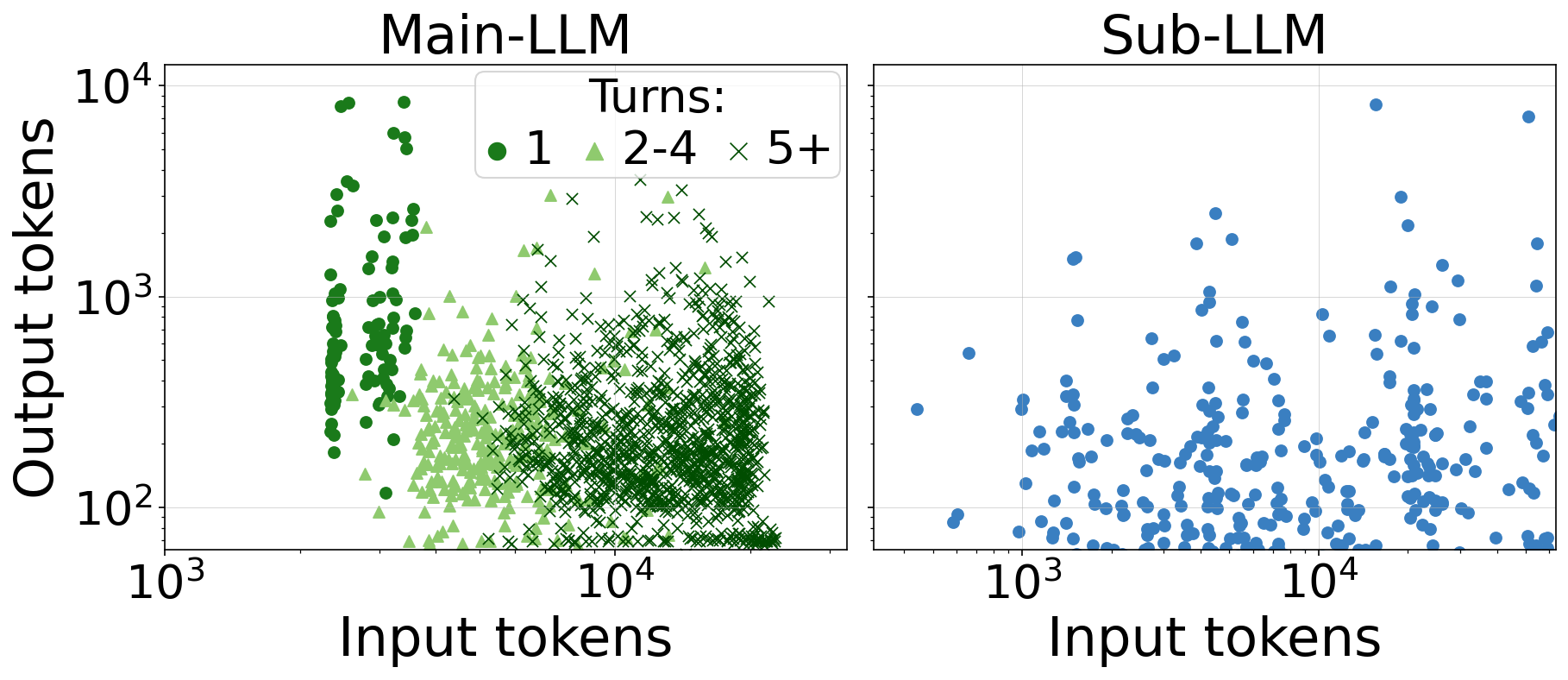}
\caption{Token distribution (prefill vs. decode) of LLMs in \mirothinker, shown in log scale. }
\label{fig:miroflow_scatter}
\end{figure}

\begin{figure}[t]
\includegraphics[width=\columnwidth]{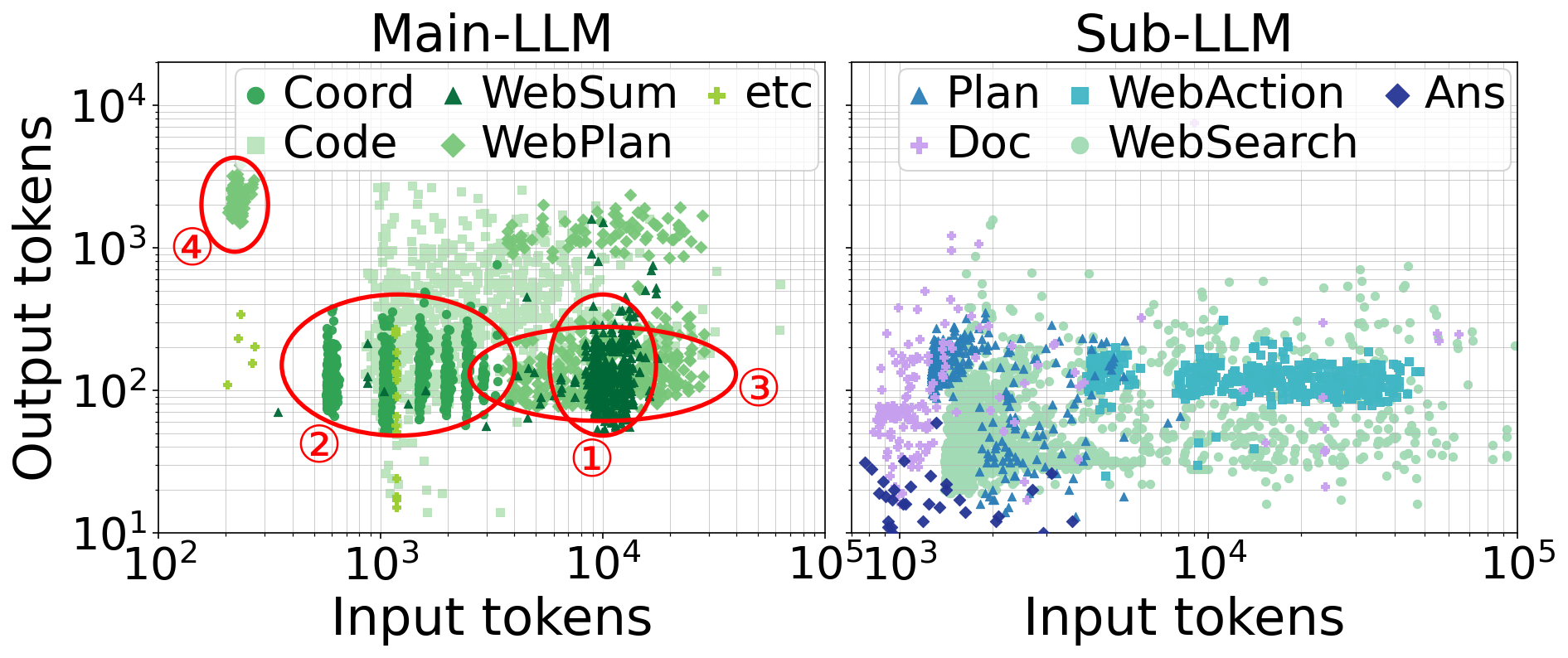}
\caption{Token distribution (prefill vs. decode) of LLMs in \owl, shown in log scale. Even the same LLM shows different prefill/decode patterns (\textit{e.g.}, \textcircled{1}--\textcircled{4}), depending on which role it is playing.}
\label{fig:owl_scatter}
\end{figure}

Next, we dissect \dataset in more detail.
Figures~\ref{fig:miroflow_scatter} and~\ref{fig:owl_scatter} show the prefill (input) and decode (output) token length distributions of queries to each LLM.
\mirothinker has a more simplistic design than \owl, and its token distributions are correspondingly simpler.
\mirothinker's \mainllm (Figure~\ref{fig:miroflow_scatter}, left) sees input tokens grow with the number of turns due to its ReAct-like design.
The first decode (Turn 1) has a notably higher average number of tokens, which is due to the initial planning step.
\mirothinker's \subllm (Figure~\ref{fig:miroflow_scatter}, right) mainly performs text summarization and thus has higher prefill tokens than decode.

In contrast, \owl exhibits more interesting behavior. Its \mainllm (Figure~\ref{fig:owl_scatter}, left) shows distinct patterns depending on which role it is playing. For example, \textcircled{1} when serving as a web summarizer, it is prefill-heavy; \textcircled{2} when serving as a coordinator, prefill tokens grow with the number of turns in ReAct-style~\cite{yao_2023_react}; and when serving as a web planner, it shows a bimodal behavior---\textcircled{3} initial planning is observation-based and prefill-heavy, while \textcircled{4} replanning (after a failure) involves deeper reasoning and is decode-heavy.
Similarly, \owl's \subllm (Figure~\ref{fig:owl_scatter}, right) also shows heterogeneous behaviors based on its role.

%\dhk{Todo: short failure trace analysis here.}
%\kwm{TODO: NEEDS WORK}

\subsection{Visualization of Representative Traces}
\label{subsec:RepresentativeTasks}

\begin{figure}[t]
\includegraphics[width=\columnwidth]{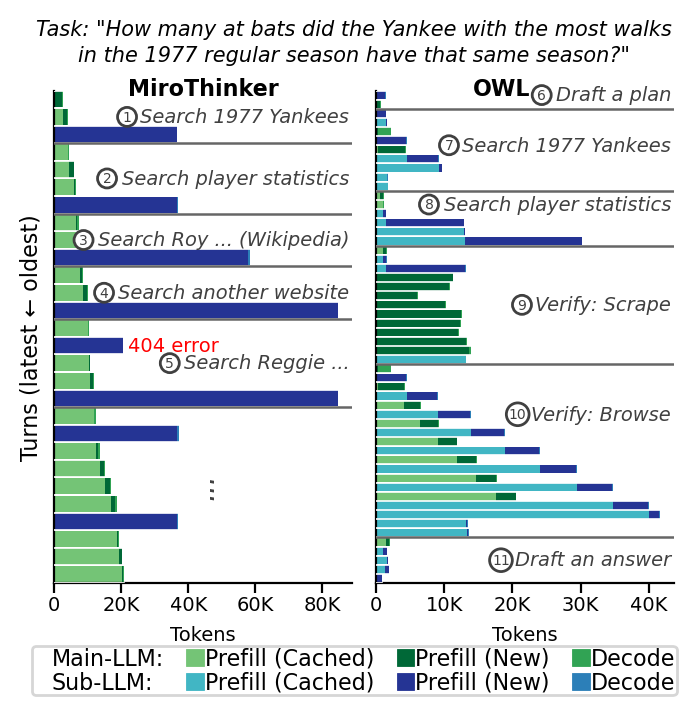}
\caption{Example traces of \mirothinker and \owl. Older queries are at the top, and newer queries at the bottom.}
\label{fig:representative}
\end{figure}

%visualizes example traces of \mirothinker and \owl on one representative task to illustrate their behavior.

To provide deeper insight into what the traces look like, we visualize example traces from \dataset in Figure~\ref{fig:representative}.
The example shows both systems answering the following question: \emph{``How many at bats did the Yankee with the most walks in the 1977 regular season have that same season?''}
Each bar represents an LLM query---\mirothinker performs 20 \mainllm and 8 \subllm inferences, while \owl performs 24 \mainllm and 30 \subllm inferences---and bar colors indicate the number of prefill/decode tokens processed and the number of prefill tokens reused from the KV cache.

In \mirothinker (left), being a ReAct-like system, the \mainllm iterates through a loop of planning, tool invocation, and observation.
Since this question requires heavy web search, only Google search and web scraping tools are exercised in this example. Each web scrape is followed by a \subllm inference (tall blue bars), which is prefill-heavy and have almost no KV cache hits.
The trace is briefly annotated in the figure.
\mirothinker sequentially \textcircled{1} searched for the 1977 Yankee roster, \textcircled{2} their at-bat and walk statistics, and \textcircled{3}--\textcircled{5} detailed information of each player, switching websites or retrying with a different URL when it encounters an error (\emph{e.g.}, a 404) or an unsatisfactory result.
The overall behavior resembles ReAct~\cite{yao_2023_react}, but does not exactly match the monotonically-increasing-prefill pattern reported by prior work~\cite{kim_2026_cost}, due to its multi-model design.

In contrast, \owl solved the problem in a more hierarchical fashion.
The coordinator \textcircled{6} first drafted an overarching plan, and specialized sub-agents executed each sub-plan.
In this example, \owl \textcircled{7} searched for the 1977 Yankees roster, \textcircled{8} pulled relevant statistics, \textcircled{9}--\textcircled{10} performed a series of verification steps, and \textcircled{11} finally drafted the answer.
Searching (\textcircled{7}--\textcircled{8}) involved web browsing and scraping, producing spikes in \subllm prefill.
\owl also attempted two distinct verification strategies.
The first (\textcircled{9}) relied on web-scraped data, incurring heavy \mainllm prefill with KV cache misses (\owl uses \mainllm when very long web data needs to be summarized, breaking them into chunks~\cite{hu_2025_owl}). When this approach failed to produce a satisfactory result, \owl fell back to a second strategy (\textcircled{10}): browsing the web interactively (processing the rendered page and scrolling down like a human), which produced alternating inferences between \mainllm (observe and plan) and \subllm (issue action).
Clearly, the two systems behave very differently even for the same task, and even conceptually similar sub-tasks (\emph{e.g.}, web search) can exhibit markedly different system behaviors (\emph{e.g.}, \textcircled{3} vs.  \textcircled{9} vs. \textcircled{10}) depending on how they are implemented.

\begin{figure}[t]
\includegraphics[width=\columnwidth]{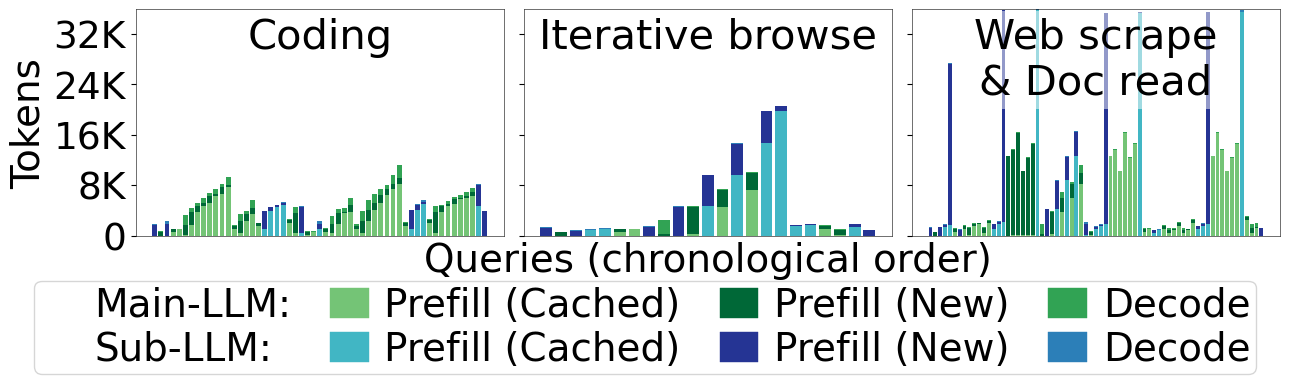}
\caption{
Additional example traces from \dataset for \owl. Each bar represents a query in chronological order.}
\label{fig:other_samples}
\end{figure}

Figure~\ref{fig:other_samples} shows additional example traces from \dataset. As the figures illustrate, behavior varies significantly across tasks.
A simple coding task (left) exhibits the ReAct-like pattern observed in prior work~\cite{kim_2026_cost}---monotonically increasing prefill with a high KV cache hit rate.
In contrast, tasks requiring iterative browsing (middle) ping-pong between \mainllm and \subllm (increasing prefills, but alternating colors), as in \textcircled{10}.
Finally, tasks involving heavy web scraping and document processing (right) exhibit bursts of extremely high prefill tokens with very low KV cache hit rates.
These examples illustrate that agentic AI's behavior is far from homogeneous, and prior studies~\cite{kim_2026_cost} have captured only a limited (though important) slice of the space.

%representative task executions from OWL, each revealing distinct workload patterns. In the coding-heavy task (left), coding agent iterates through a repeated cycle of code generation and debugging, producing a stable and moderate token footprint.
%In the video-and-browse task (middle), the agent first retrieves a summarized answer from \texttt{ask\_question\_about\_video}, incurring minimal tokens, before switching to iterative web browsing (\texttt{browse\_url}) to  verify the answer. resulting in sharp token spikes as highlighted.
%In web search task (right), the agent issues multiple searches and document extractions across different sources. 
%\owl routes document extraction through two models depending on content length (shown in Figure~\ref{fig:overview}), by splitting the content and manifests as bursts of large prefill tokens during scraping. Here, browse action and eval interleaved amplifying token variability (\texttt{browse\_url}).

%We summarize our observations below:

\paragraph{\textbf{Takeaway 1}}
\dataset emphasizes that complex agentic AI systems' behavior is highly heterogeneous: the statistics have extremely high variance, and behavioral patterns vary between agent architecture designs, agent roles, models, and tasks.
%, and prefill tokens do not always monotonically increase as in simple ReAct-like systems~\cite{cost_of_dynamic_reasoning, react}.
%
Even the same LLM experiences different prefill-decode behaviors depending on the overall system design (\textit{e.g.}, whether it is ReAct-like), which agent role it serves as (\textit{e.g.}, coordinator vs. summarizer), and other surrounding contexts (\textit{e.g.}, initial planning vs.\ replanning).
Our results suggest that system optimizations tuned for simple agentic designs and workloads may not transfer to the more complex designs and workloads of the future, motivating further study of these more complex setups.

%indicate that using a static and global system configuration might be suboptimal, and task-, model-, and agent-aware design need to be studied. 
%
%Also, accounting for reasoning tokens is critical for accurate performance modeling, which \dataset fully includes.

\subsection{Tool Statistics}
\label{subsec:toolanalysis}

\begin{table}[t]
\centering
\caption{Per-tool execution statistics.
Times are in seconds.
%Tools with $^\dagger$ are followed by summarization with \subllm.
%\dhk{$^\dagger$ are tools with only scraping execution time, while LLM-assisted extraction(summarization) replayed in \subllm.} \kwm{What is this note with $^\dagger$ here?}
}
\label{tab:tool_stats}
\resizebox{\columnwidth}{!}{%
\begin{tabular}{llrrrr}
\toprule
\textbf{} & \textbf{Tools} & \textbf{Calls} &
\textbf{Min} & \textbf{Max} & \textbf{Mean±std}  \\
\midrule
\multirow{5}{*}{\rotatebox[origin=c]{90}{\textbf{\mirothinker}}}
 & \texttt{google\_search}                 & 
 539 & 0.56 & 9.18 & 1.85±1.23\\
 & \texttt{scrape\_and\_extract\_info}               & 622 & 0.33 & 422.36 & 3.70±15.46\\
 & \texttt{run\_python\_code}                          & 136 & 0.47 & 32.96 & 1.88±2.82 \\
 & \texttt{create\_sandbox}    &  19 &0.92 &1.98 &1.10±0.18 \\
 \midrule
\multirow{10}{*}{\rotatebox[origin=c]{90}{\textbf{\owl}}}
 & \texttt{search\_google}             & 686 & 0.14  & 5.30  &  0.30±0.15\\
 & \texttt{extract\_document\_content}            &419  & 0.00  &  91.08  &  13.78±10.61\\
 & \texttt{execute\_code}                      & 262 &  0.02 &  60.07  &  1.48±6.92\\
 & \texttt{search\_wiki}                        & 166  & 0.09  &  1.57  &  0.48±0.24\\
 & \texttt{search\_wiki\_revisions}               & 84 &  0.07 &   0.29 &   0.10±0.03 \\
 & \texttt{extract\_excel\_content}               & 36 &  0.00 &   1.05 &    0.05±0.14 \\
 & \texttt{ask\_question\_about\_image}             & 19 & 0.05 &  20.59  &  4.61±5.62 \\
 & \texttt{search\_archived\_webpage}              & 11 & 0.25  & 10.62  &  2.92±3.77    \\
 & \texttt{ask\_question\_about\_video}             & 12 &   2.60 &  42.61  & 14.74±12.03  \\
 & \texttt{ask\_question\_about\_audio}             & 5 & 1.33  &  4.04  &  2.44±0.84 \\
\bottomrule
\end{tabular}}
\end{table}

Table~\ref{tab:tool_stats} summarizes tool call frequencies and their execution time statistics.
Tools fall into roughly three categories.
First, fast tools (\texttt{search\_google}, \texttt{search\_wiki}, \texttt{extract\_excel\_content}, \textit{etc.}) complete within seconds and do not noticeably affect end-to-end latency.
Second, tools like \texttt{extract\_document\_content} rely on external webscraper (\textit{e.g.} Jina~\cite{jina.ai} and Firecrawl~\cite{firecrawl}) and exhibits much higher latency due to the website or the service being slow. These contribute non-negligibly to end-to-end latency, but little can be done about it.
Finally, LLM-assisted tools (\texttt{ask\_question\_about\_video}, \texttt{ask\_question\_about\_audio}) use proprietary LLMs to handle inputs that \mainllm/\subllm cannot process (\textit{e.g.}, YouTube videos are processed by Google's Gemini-2.5 via API).
We currently model them as black-box tools, but they could potentially be replaced with local models (another \subllm) given additional engineering effort.
%and/or be accelerated by improving the LLM serving system.

\paragraph{\textbf{Takeaway 2}}
Tool latency is dominated by a small subset of tools that rely on slow external services or LLMs.
Their behavior has high variance cannot be accelerated easily.
Also, their latencies are difficult to model systematically, posing a practical challenge for trace-driven simulation studies. The issue can be potentially eased by replacing tools using proprietary LLMs with open-source LLMs, but this requires non-negligible effort.
\section{System Characterization with \sys}
\label{sec:simulation}

Next, we show results of our system characterization study performed by running \dataset on \sys under various system configurations.
The following results serve two main purposes: (1) it provides us insights on the system overheads of complex agentic AI systems, and (2) it demonstrates the potential use of \dataset and \sys on agentic AI system research.

%\subsection{Experimental Setup} 
%\label{subsec:experimental_setup}
%We adopt Vidur~\cite{vidur}, an LLM inference simulator that
%estimates end-to-end latency through a tree execution time predictor based on profiled result.
%We extend Vidur with three contributions: 
%(1) Multi-turn request support with inter-request tool call latency 
%modeling, enabling simulation of full agentic task traces.
%(2) Heterogeneous model configurations including prefill--decode (PD) 
%disaggregation, allowing prefill and decode phases to be assigned to 
%separate GPU partitions with configurable TP degrees. and 
%(3) Additional scheduling policies beyond FCFS, including 
%prefix-cache-aware and shortest-remaining-prefill-first variants, 
%detailed in Section~\ref{subsec:scheduling}.

\subsection{Simulation Setup}
\label{sec:simulation_setup}

We simulated a cluster consisting of two DGX A100 nodes (16$\times$A100 (80GB) GPUs).
By default, we ran four models (two \mainllms and \subllms), each with tensor parallelism degree of four (TP-4), and adopted prefill-decode disaggregation. That is, the default setup ran \mainllm prefill, \mainllm decode, \subllm prefill, and \subllm decode on four GPUs (TP-4) each.
Section~\ref{subsec:modelconfig} varies this configuration.
Since Vidur does not yet ship a performance model for the exact LLMs we used, we used Qwen2.5-32B-Instruct's~\cite{qwen25_2025} performance model instead.
We expect this to still capture a reasonable first-order estimate of the overall trends, as Qwen2.5 shares its core architectural features---a decoder-only Transformer with grouped-query attention (GQA)~\cite{ainslie_2023_gqa} and rotary position embeddings (RoPE)~\cite{su_2023_rope}---with many popular LLMs. A performance model fitted to our exact LLMs would tighten the absolute numbers.
%We expect this to still capture a relative first-order estimate of the trend, as CodeLlama shares the same decoder-only Transformer structure with grouped query attention (GQA)~\cite{ainslie_2023_gqa} and RoPE~\cite{su_2023_rope} with many popular LLMs; a performance model fitted to our exact LLMs would tighten the absolute numbers.
%, which popular models (e.g. Qwen2.5, Llama3) are already using.
%
%Our setup can be seen as simulating a scenario where a CodeLlama-like model generates similar behavioral traces.
%
%While not exact, we believe the simulation result will preserve relative trends and serve as a reasonable first-order analysis; 
%
%Generating additional performance models is a capability Vidur~\cite{vidur} already provides.
In Section~\ref{subsec:scalability}, we increase the size to Llama-3-70B.

We simulate the task arrival with Poisson distribution with varying rate (0.05--0.5 tasks per second, with 0.1 tasks per second being the default). Note that we control the tasks per second (TPS), not queries per second (QPS).
Each task invokes multiple queries, so TPS is generally lower than QPS. Incoming tasks are randomly shuffled, and prefix caching is enabled by default, except for in Section~\ref{subsec:prefixcache} where it is varied.
% \kwm{@Donghwan: How large is the KV cache?}
%\dhk{Following VLLM V1~\ref{todo}, the KV cache is maintained by block-granularity (16 tokens), while Each TP-4 GPUs can hold 32760/\# blocks in 35/70B model, respectively.}
Following vLLM V1~\cite{vllm_v1, kwon_2023_efficient}, KV cache is managed at block granularity (16 tokens per block); each TP-4 group holds roughly 58{,}848 blocks for the 32B model.
and 32{,}870 blocks for the 70B model.

%\subsection{Prefill Overhead in Agentic Workloads}

%Although prefill benefits from token-level parallelism, it holds exclusive use of the prefill replica for its entire duration.
%Unlike decode, where multiple requests are batched and make progress simultaneously, a single long prefill blocks every other queued request from being scheduled until it completes.
%The resulting queuing delay is proportional to the prefill length of the head-of-line request, not the number of waiting sessions, making long prefills disproportionately harmful to overall throughput.

%Agentic workloads systematically produce long and uncached prefills.
%Each agent turn ingest previously unseen content like retrieved web pages, scraped documents, tool outputs, and multi-modal inputs.
%Beyond penalizing the affected request directly, these cache-miss prefills rapidly fill the KV cache pool, evicting retained prefix blocks from prior turns and degrading the hit ratio for subsequent requests toward the no-cache baseline.

%We evaluate five dimensions of this problem: prefix caching
%(Section~\ref{subsec:prefixcache}), scheduling policy
%(Section~\ref{subsec:scheduling}), model configuration
%(Section~\ref{subsec:modelconfig}), KV cache eviction and recomputation
%(Section~\ref{subsec:eviction}), and chunked prefill vs.\ PD disaggregation
%(Section~\ref{subsec:pdvschunked}).

\begin{figure}[t]
\includegraphics[width=\columnwidth]{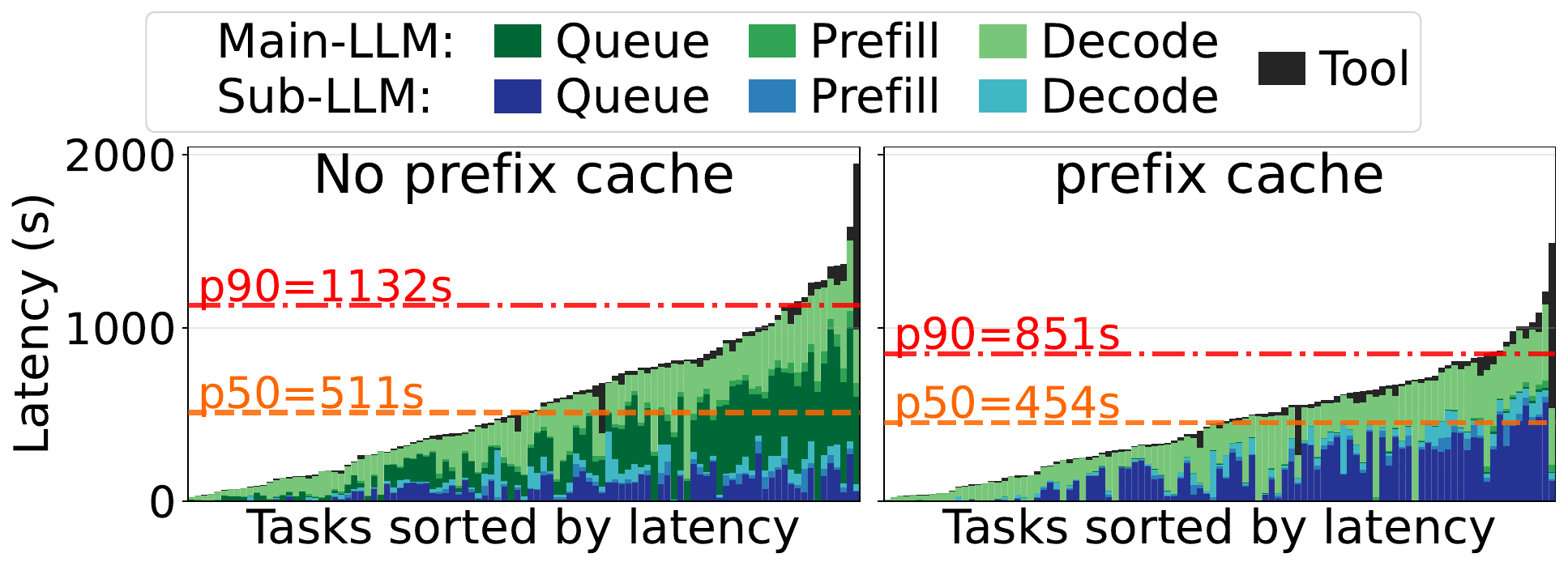}
\caption{Effect of prefix caching in \mirothinker. Each task is sorted by its end-to-end latency.}
\label{fig:prefix_caching_miro}
\end{figure}

\begin{figure}[t]
\includegraphics[width=\columnwidth]{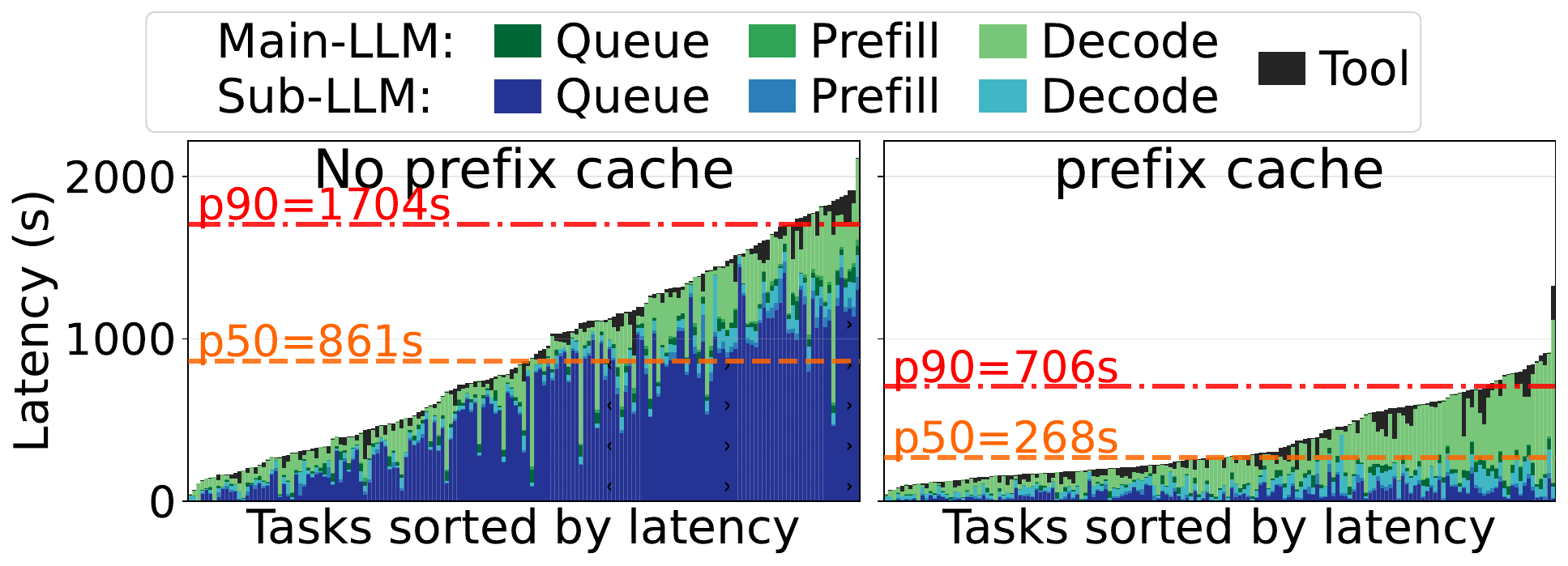}
\caption{Effect of prefix caching in \owl. Each session sorted by its end-to-end latency. }
\label{fig:prefix_caching_owl}
\end{figure}

\subsection{Initial Overhead Analysis}
\label{subsec:overheads}

First, we studied the overheads of prefill, decode, and tool calls.
Figures~\ref{fig:prefix_caching_miro} and~\ref{fig:prefix_caching_owl} show the latency of \mirothinker and \owl tasks with and without prefix caching, sorted by end-to-end latency.
We break down the overheads into \mainllm and \subllm's prefill, decode, queuing delay at the prefill GPUs, and tool call overhead. Decode, unlike prefill, can be batched and run in parallel, so queuing delay at decode GPUs was negligible and omitted.

First, the bottleneck varied with the system.
Without prefix caching, \mirothinker was bottlenecked by \mainllm prefill, whereas \owl was bottlenecked by \subllm prefill.
This is because \mirothinker's \mainllm is ReAct-like, so prefill tokens accumulate quickly over turns, and {\owl's \subllm frequently processes long text or images, incurring larger prefill overheads.}
Second, unlike prior reports~\cite{kim_2026_cost}, tool calls were not a major overhead, because GAIA tasks require heavy text/image ingestion and long reasoning, making LLM overheads dominant. Still, extremely slow web services occasionally drove up end-to-end latency.
Finally, decode of \mainllm (dominated by reasoning) and prefill of \subllm together accounted for a significant portion of the overall overhead (especially with prefix caching), neither of which is captured by existing datasets such as MiroVerse~\cite{miroverse}. This underscores the value of \dataset.

\paragraph{\textbf{Takeaway 3}}
System overheads depend on agent design, and our findings do not simply replicate prior observations~\cite{kim_2026_cost}. For complex tasks like GAIA, LLM overheads dominated in most cases and tool call overheads were small, though slow external services occasionally drove up end-to-end latency.
Moreover, reasoning tokens and \subllm activity, both not captured by existing trace datasets, accounted for a sizable share of the total overhead.

\subsection{Impact of Prefix Caching}
\label{subsec:prefixcache}

Figures~\ref{fig:prefix_caching_miro} and~\ref{fig:prefix_caching_owl} show that prefix caching significantly improves both systems: p50 end-to-end latency improves by 1.13$\times$ (\mirothinker) to 3.21$\times$ (\owl), and p90 by 1.33$\times$ (\mirothinker) to 2.41$\times$ (\owl).
These are larger than what prior study~\cite{kim_2026_cost} reported (15.7\% end-to-end improvement~\cite{kim_2026_cost}), especially for OWL, for two main reasons.
First, GAIA's complexity demands huge ingestion of searched data, making workloads more prefill-heavy.
Second, we simulate batched inference, whereas prior work~\cite{kim_2026_cost} studied single-batch inference. Prefill benefits less from batching as it is compute-bound, so its share of latency is even larger in our setting.
Consequently, prefill (compute plus queuing) dominates without prefix caching (Figures~\ref{fig:prefix_caching_miro} and~\ref{fig:prefix_caching_owl}, left), and its large overhead shrinks substantially once prefix caching is enabled (right).
\mirothinker's prefill remains large even with prefix caching because its \subllm processes long prefills with an extremely low KV-cache hit rate (Table~\ref{tab:trace_stats}); \owl, in contrast, becomes decode-bound.

\begin{figure}[t]
\includegraphics[width=\columnwidth]{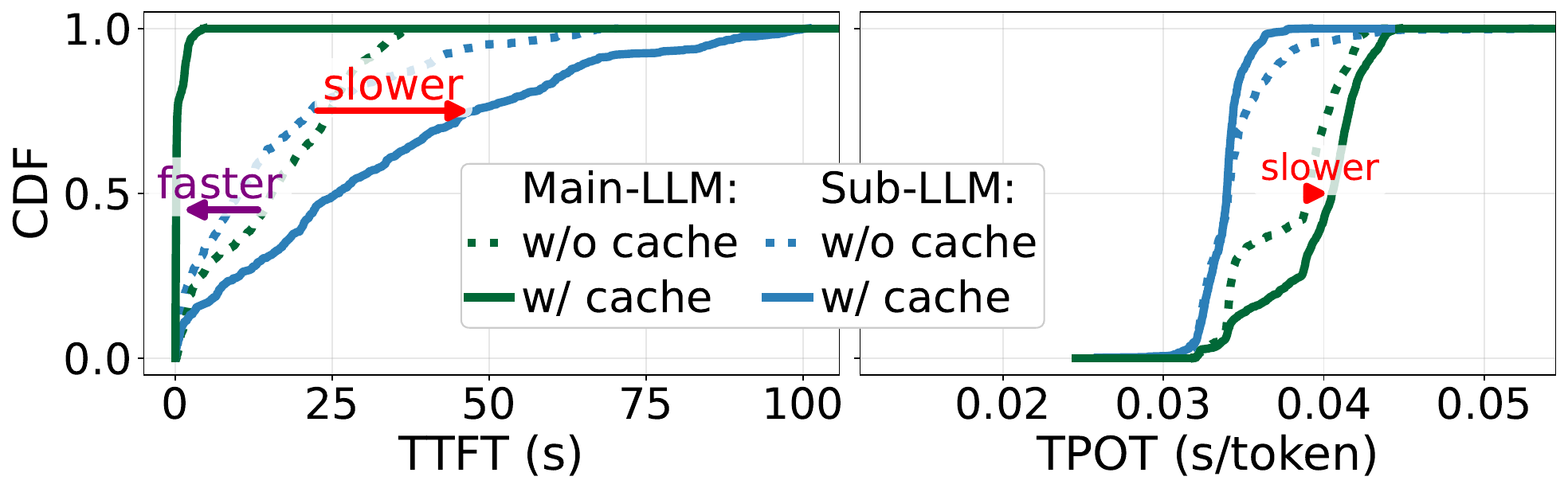}
\caption{Cumulative distribution function (CDF) of \mirothinker's TTFT and TPOT with and without prefix caching.
}
\label{fig:prefix_caching_miro_ttft_tpot}
\end{figure}

\begin{figure}[t]
\includegraphics[width=\columnwidth]{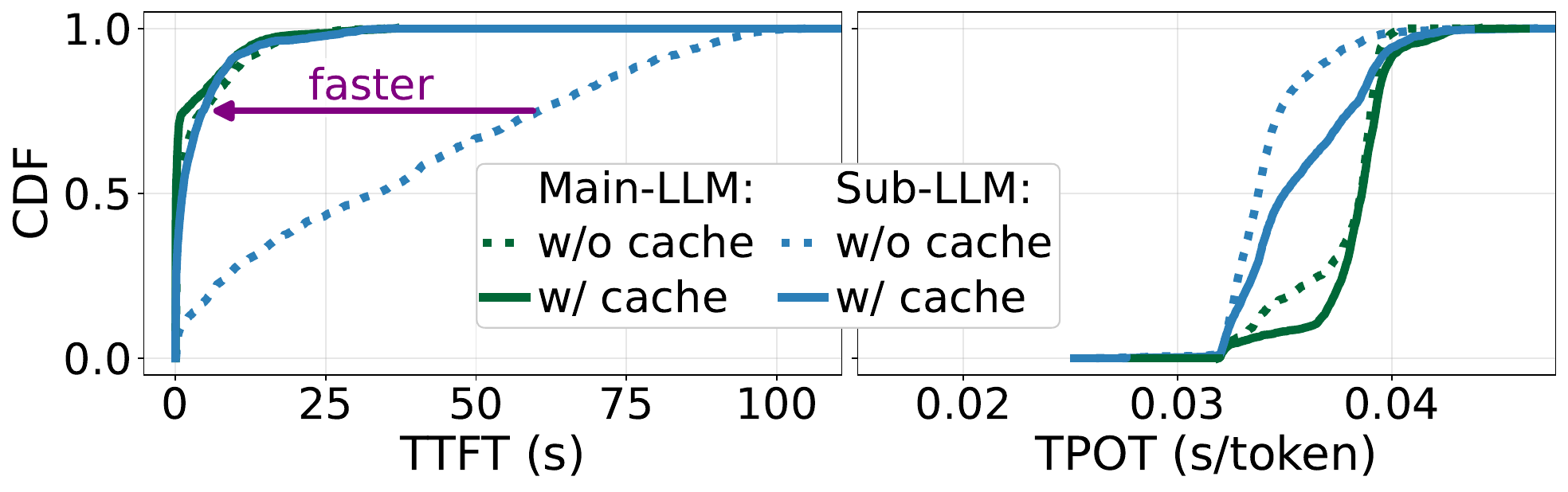}
\caption{Cumulative distribution function (CDF) of \owl's TTFT and TPOT with and without prefix caching. }
\label{fig:prefix_caching_owl_ttft_tpot}
\end{figure}

Figures~\ref{fig:prefix_caching_miro_ttft_tpot} and~\ref{fig:prefix_caching_owl_ttft_tpot} further break down how time-to-first-token (TTFT) and time-per-output-token (TPOT) change with prefix caching.
Figure~\ref{fig:prefix_caching_miro_ttft_tpot} (left) shows that, in \mirothinker, the \mainllm's TTFT improves (CDF shifts left) as expected, but \subllm's TTFT \emph{degrades} (CDF shifts right), which is counterintuitive.
This happens because: (1) \mirothinker's \subllm has an extremely low KV-cache hit rate (Table~\ref{tab:trace_stats}) and barely benefits from prefix caching, and (2) as the \mainllm benefits from prefix caching, it submits work to \subllm more frequently, lengthening its prefill queue.
TPOT increases, but only slightly, because decode runs at a larger batch.
%size once prefill clears faster.

Unlike chatbot applications where individual TTFT and TPOT are important service-level objective (SLO), the \textit{task completion latency} is the most important metric in agentic AI systems.
End-to-end task latency still improves overall, since the \mainllm's TTFT speedup dominates.
Figure~\ref{fig:prefix_caching_owl_ttft_tpot} shows a similar trend for \owl---its \subllm's TTFT significantly improves---except its \mainllm's TTFT does not noticeably degrade. This is because \owl's \mainllm also enjoys a reasonable KV cache hit rate.
%\kwm{Something's weird. Why is Figure 6 and 8 showing different bottleneck?}

\paragraph{\textbf{Takeaway 4}}
Prefix caching yields larger gains in our setup than in prior reports~\cite{kim_2026_cost}, but not all components benefit: while \mainllm's TTFT improved substantially, some metrics worsen slightly---or significantly, as with \mirothinker's \subllm TTFT.
This is a new observation that was not reported in previous works that focused on rather simplistic agentic AI systems (\textit{e.g.}, ReAct~\cite{yao_2023_react, kim_2026_cost}), and implies that the effect of prefix caching is model- and system-specific.
When considering the end-to-end task latency, it improved with prefix caching as expected.
%
%However, unlike user-facing 

\subsection{Impact of Task Arrival Rate}
\label{sec:arrival_rate}

\begin{figure}[t]
\includegraphics[width=\columnwidth]{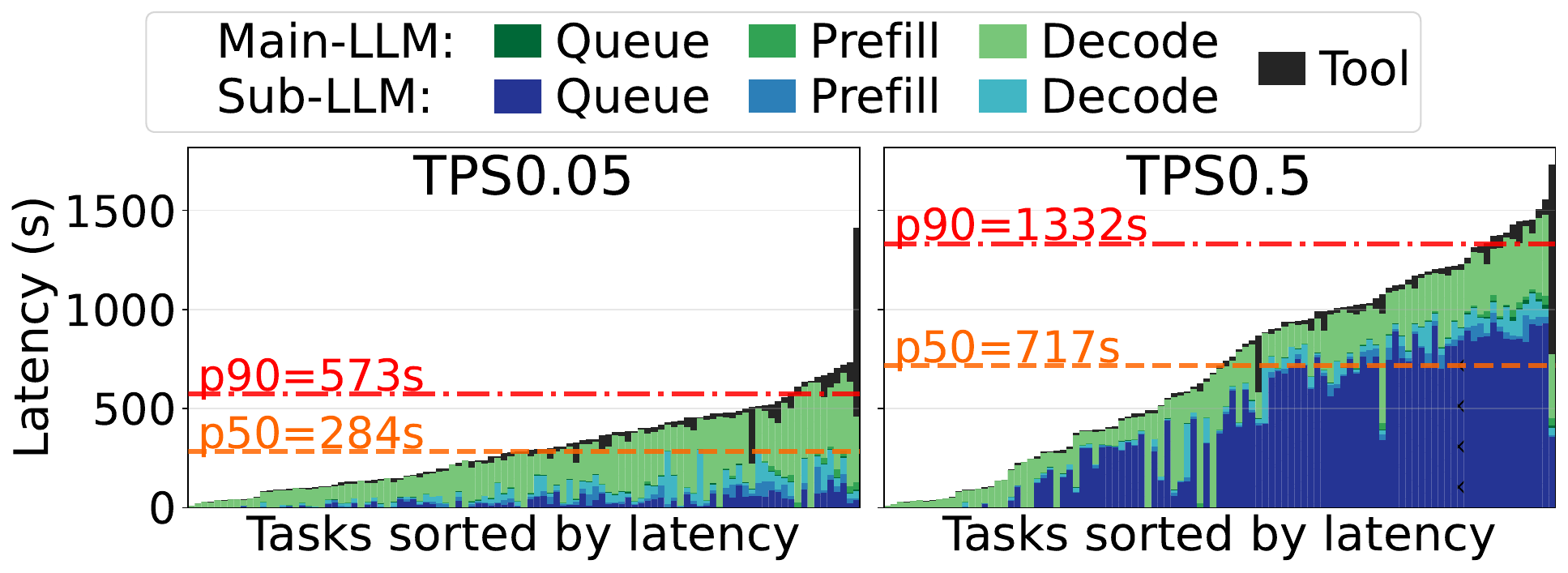}
\caption{Task latency with different TPS in \mirothinker.
}
\label{fig:miro_tps}
\end{figure}

\begin{figure}[t]
\includegraphics[width=\columnwidth]{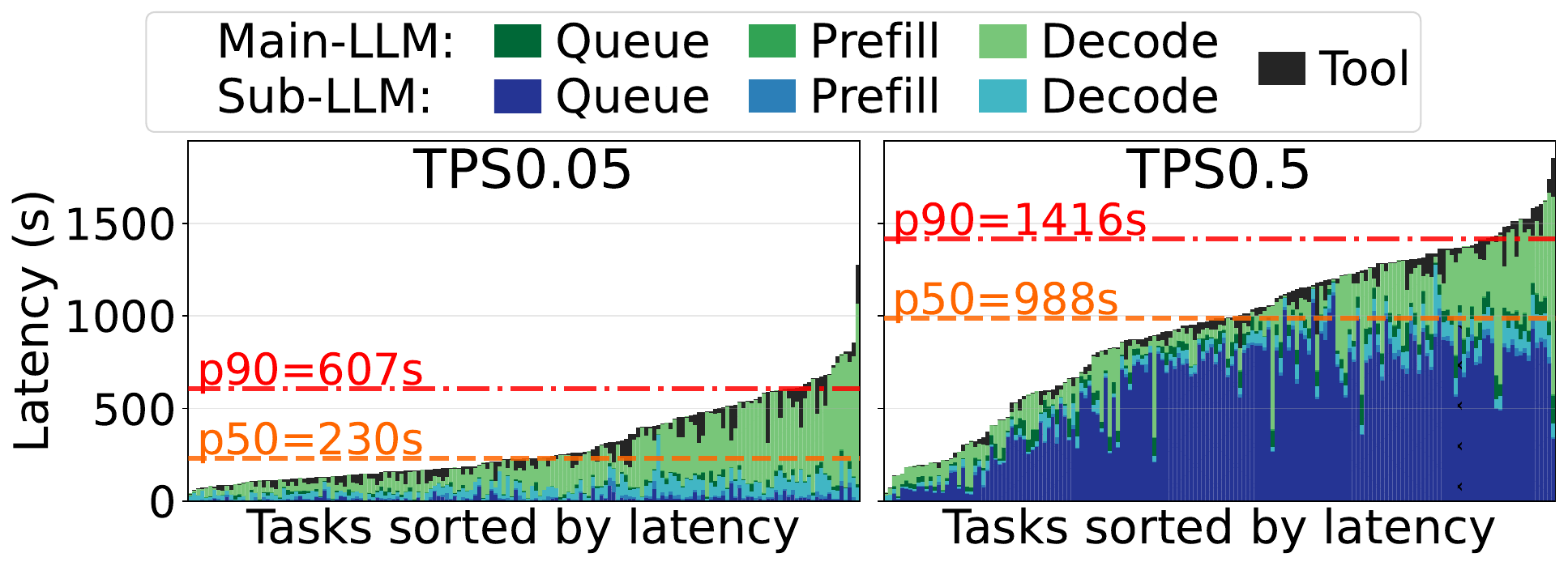}
\caption{Task latency with different TPS in \owl.}

\label{fig:owl_tps}
\end{figure}

Figures~\ref{fig:miro_tps} and~\ref{fig:owl_tps} show how end-to-end task latency changes with the task arrival rate (tasks per second, or TPS).
Compared to TPS=0.1 (Figures~\ref{fig:prefix_caching_miro} and \ref{fig:prefix_caching_owl}, right), per-task completion time decreased at lower TPS (0.05) and increased at higher TPS (0.5), which is expected as the system becomes more bottlenecked.
The bottleneck itself also shifted with TPS. At low TPS (0.05), the bottleneck was mainly the \mainllm's decode, while at high TPS (0.5), the \subllm's prefill became the bottleneck. This is again expected because prefill is less batch-able than decode.

\paragraph{\textbf{Takeaway 5}}
At low arrival rates, the reasoning model's (\mainllm) decode was the main bottleneck. At higher arrival rates, prefill capacity dominated.
This indicates that the bottleneck varies with the arrival rate, and the optimal configuration may shift accordingly.

\subsection{Impact of Scheduling Policy} 
\label{subsec:scheduling}

Next, we studied the impact of query scheduling policies.
Scheduling policies primarily affect queuing delay, so we focused on the setup where queuing delay was most dominant: TPS=0.5. The trend was similar at lower TPS, but less pronounced.
We also restricted our study to scheduling policies on the prefill GPUs, since decode GPUs do not exhibit notable queuing delay.
Prefill scheduling matters because a single long prefill can occupy the entire prefill GPUs and block every other query behind it, so the order in which queries are dequeued directly shapes tail latency.
We explore the four policies listed below; all previous figures used the simplest, Q-FCFS.
\begin{enumerate}[topsep=5pt, itemsep=0pt, leftmargin=*]
    \item \textbf{Query-level first-come-first-serve (Q-FCFS):} Queries are served in the order they arrive in the prefill queue.
    \item \textbf{Task-level first-come-first-serve (T-FCFS):} Queries are prioritized by the arrival time of the task they belong to, so all queries from an earlier task are served before any query from a later task~\cite{li_2026_continuum}.
    \item \textbf{Shortest-job-first (SJF)}~\cite{linus_1968_sjf}: Queries with the shortest remaining prefill are prioritized to reduce head-of-line blocking. Though non-clairvoyant decoding limits SJF elsewhere~\cite{wu_2023_fast, luo2025autellix}, GAIATrace’s huge prefills and short decodes perfectly enable this approach.
    \item \textbf{SJF with timeout}: Same as SJF, but any query that has waited longer than a threshold is promoted to high priority, preventing starvation. 
\end{enumerate}

%\begin{figure}[t]
%\includegraphics[width=\columnwidth]{Figures/gantt_Miro4096_type2.png}
%\caption{End-to-end task latency with different scheduling policy in Miroflow, low task arrival time (QPS=0.05).
%\dhk{Might be deleted}
%}
%\label{fig:miro_scheduling_low}
%\end{figure}

\begin{figure}[t]
\includegraphics[width=\columnwidth]{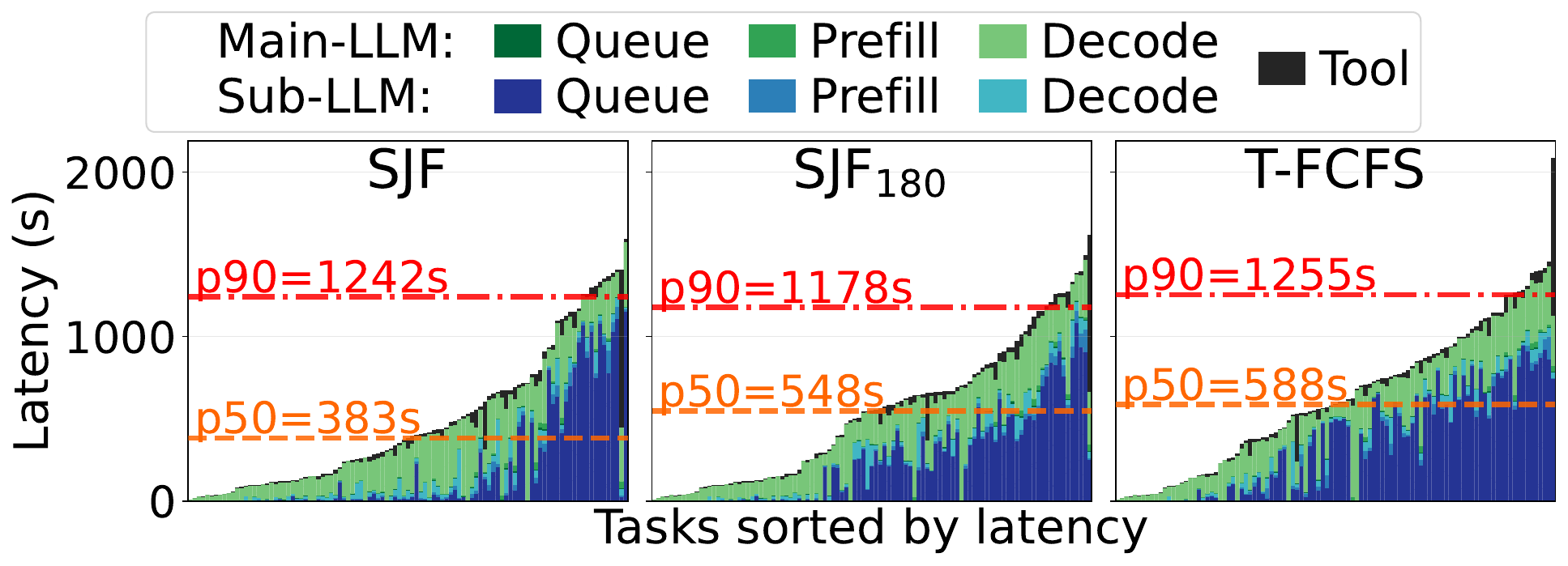}
\caption{Different scheduling policies (\mirothinker, TPS=0.5; Figure~\ref{fig:miro_tps} (right) is Q-FCFS for the same setup).
%\dhk{need TTFT TPOT for different HW config}
%\kwm{only leave SJF inf and SJF180. Add T-FCFS}
}
\label{fig:miro_scheduling_high}
\end{figure}

\begin{figure}[t]
\includegraphics[width=\columnwidth]{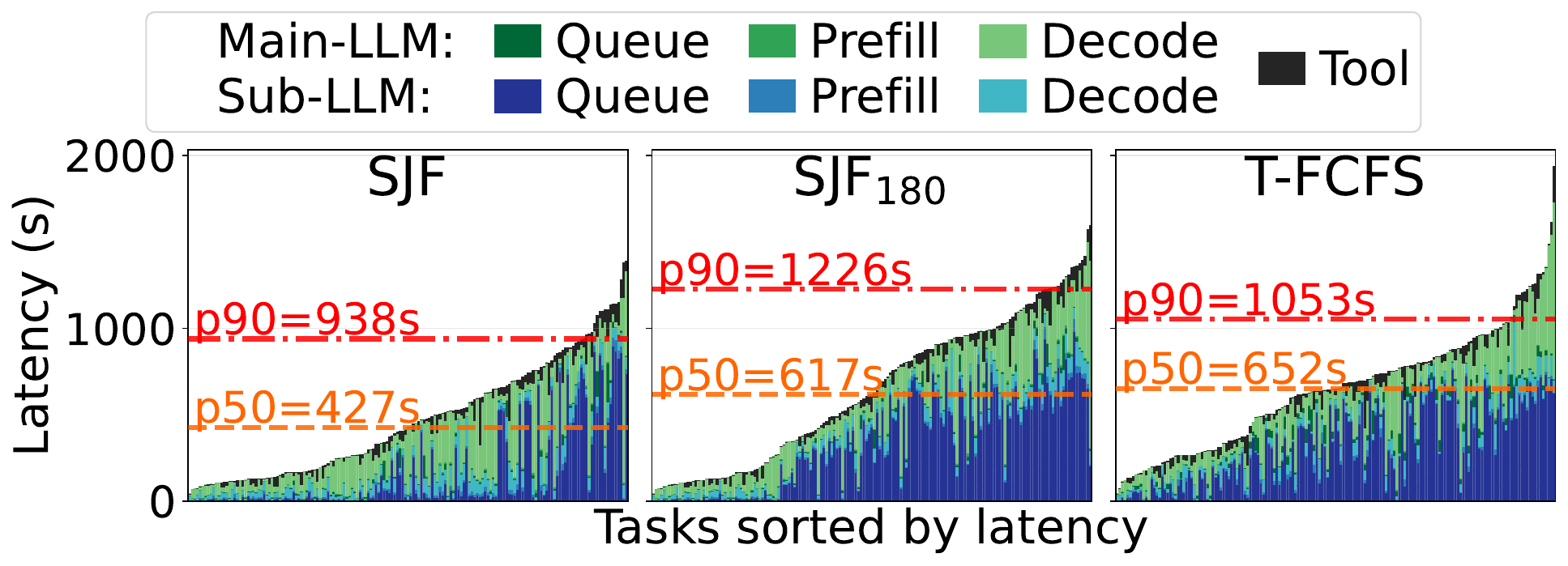}
\caption{Different scheduling policies (\owl, TPS=0.5; Figure~\ref{fig:owl_tps} (right) is Q-FCFS for the same setup).
}
\label{fig:owl_scheduling_high}
\end{figure}

%(Figures~\ref{fig:miro_scheduling} and \ref{fig:owl_scheduling})

Figures~\ref{fig:miro_scheduling_high} and~\ref{fig:owl_scheduling_high} show the task latencies when using other scheduling policies.
Compared to Q-FCFS (Figures~\ref{fig:miro_tps} and~\ref{fig:owl_tps}, right), SJF significantly improved overall end-to-end latency, reducing p50 latency by 1.87$\times$ (\mirothinker) and 2.31$\times$ (\owl).
This is because a long prefill query has two adverse effects under Q-FCFS: (1) it occupies the prefill GPUs and blocks all queries behind it (\textit{i.e.}, head-of-line blocking), and (2) its large KV cache evicts other useful entries.
By deferring longer prefills and running shorter ones first, SJF mitigates both effects.

\begin{figure}[t]
\includegraphics[width=\columnwidth]{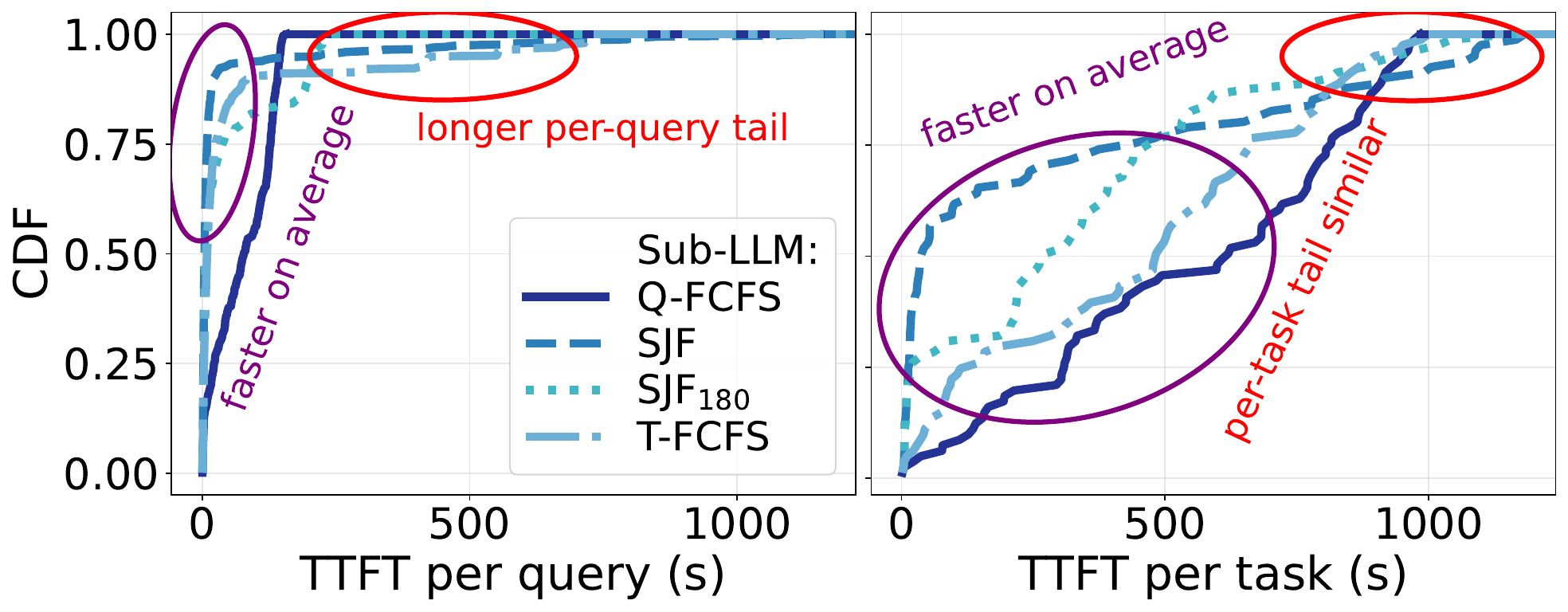}
\caption{Per-query and per-task TTFT with different scheduling policies for \subllm (TPS=0.5, \mirothinker; \owl shows similar trend and omitted). While per-query TTFT significantly worsened around the tail when using SJF (compared to the baseline Q-FCFS), per-task tail did not worsen significantly. On average (near p50), all the other scheduling policies improved upon Q-FCFS.
\label{fig:miro_req_vs_session}
}
\end{figure}

Counterintuitively, deferring long prefills did \textit{not} harm task-level tail latency---even p90 slightly improved under SJF. To understand why, we additionally plotted the CDF of per-query TTFT and per-task TTFT in Figure~\ref{fig:miro_req_vs_session} (only \subllm TTFT shown, which is the dominating bottleneck).
Figure~\ref{fig:miro_req_vs_session} (left) shows that per-query TTFT increased sharply at the tail under SJF, which is expected: queries with long prefills are repeatedly deferred and accumulate delay.
However, task-level sum of TTFT at the tail did \textit{not} worsen much (Figure~\ref{fig:miro_req_vs_session}, right).
The reason is that a task issues multiple queries, and SJF's overall reduction in queuing delay also benefits the \textit{other queries from the same task} enough to offset one query's TTFT becoming significantly worse.
%
%In other words, even if one query within a task suffers a much longer TTFT, the remaining queries from that task become significantly faster, so the task-level latency comes out ahead. 
In our experiment, SJF almost always won at the task level, although it looks harmful in per-query TTFT.

Adding a 180-second timeout ($\text{SJF}_{180}$) slightly improved the tail (\textit{e.g.}, p90) latency for \mirothinker but worsened it for \owl.
The reason is similar: promoting a long-waiting query (which is, in SJF, a long-prefill query) improves TTFT for that single query, but blocks the prefill GPUs and evicts KV cache entries for everyone else---and \textit{other queries from the same task} also hurt.
So while the timeout shortened the per-query TTFT tail (in Figure~\ref{fig:miro_req_vs_session} (left), $\text{SJF}_{180}$ has a much shorter tail than SJF), the end-to-end task latency almost always worsened with timeout in our experiments.
Task-level FCFS (T-FCFS) showed a slight improvement over Q-FCFS, because prioritizing queries from already-active tasks increases the chance of their KV cache entries being reused before eviction. The effect, however, was marginal.

\paragraph{\textbf{Takeaway 6}}
Scheduling policies that reduce prefill queuing delay, such as SJF, can significantly improve overall task latency.
Surprisingly, a longer per-query tail TTFT did not always translate to a longer per-task tail TTFT or worse end-to-end latency.
This suggests that prior optimizations targeting per-query TTFT/TPOT are not necessarily optimal for agentic AI systems, whose primary metric of interest is per-task latency, not per-query metrics.

%Once again, the per-query and per-task views disagree: for agentic AI, end-to-end task latency is the metric that matters, and optimizations targeting per-query TTFT or TPOT do not automatically translate to per-task improvements.

\subsection{Impact of System Configuration}
 \label{subsec:modelconfig}

%\begin{figure}[t]
%\includegraphics[width=\columnwidth]{Figures/gantt_Miro4096_type7.png}
%\caption{Different Hardware setup in Miroflow with moderate task arrival rate (QPS0.1) %\label{fig:miro_model_config_medium}}
%\end{figure}

We also studied the effect of varying the number of GPUs assigned to each model's prefill and decode stages.
The baseline (Balanced) assigns four GPUs (TP-4) to every prefill and decode stage, as explained in Section~\ref{sec:simulation_setup}.
We additionally studied two setups that give more GPUs to the \subllm's prefill stage, which is the dominant bottleneck at high TPS, by taking GPUs away from different stages.
The first takes GPUs from the \mainllm's decode (reducing to TP-2), and the second from the \mainllm's prefill.

\begin{figure}[t]
\includegraphics[width=\columnwidth]{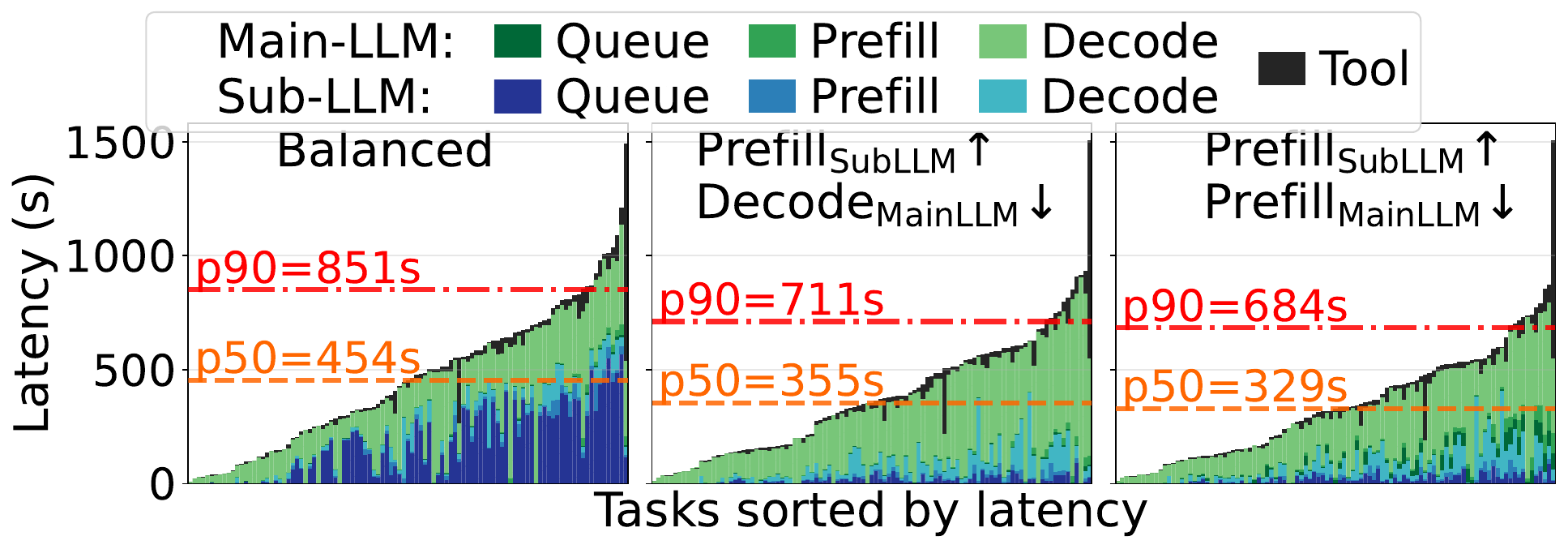}
\caption{Different system configuration in \mirothinker with medium task arrival rate (TPS=0.1).
\label{fig:miro_model_config_low}}
\end{figure}

\begin{figure}[t]
\includegraphics[width=\columnwidth]{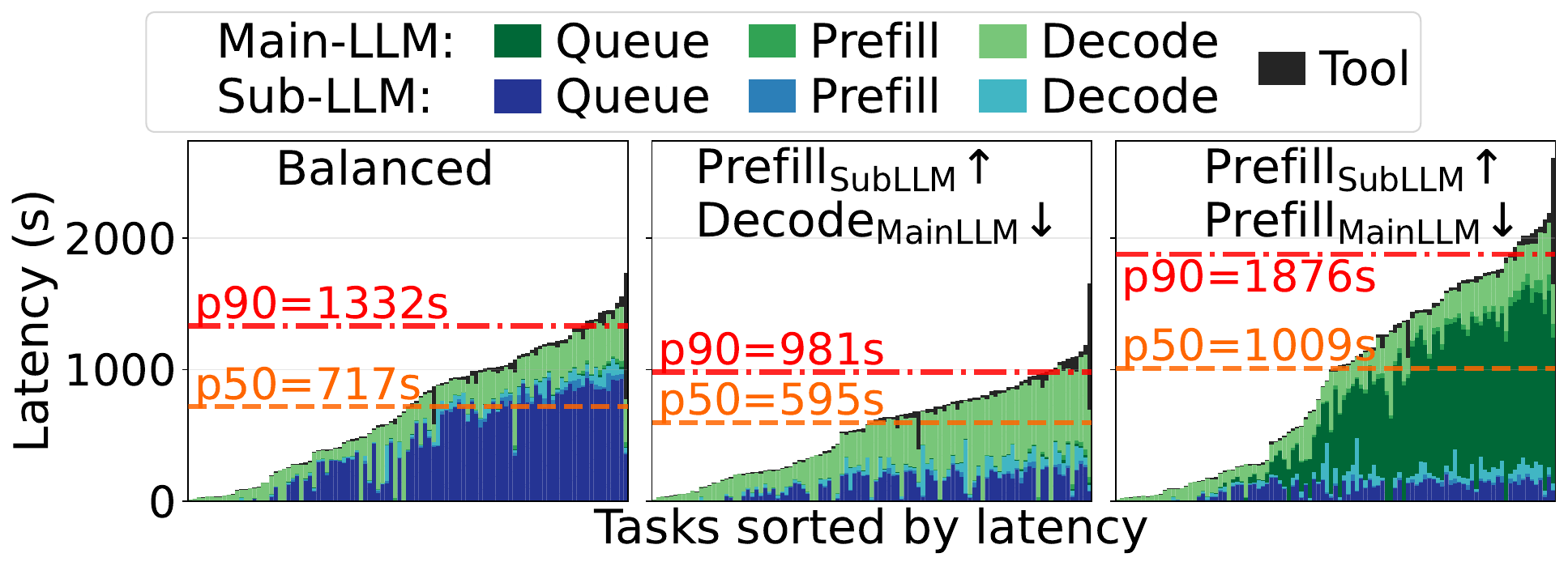}
\caption{Different system configuration in \mirothinker with high task arrival rate (TPS=0.5).
\label{fig:miro_model_config_high}}
\end{figure}

Figures~\ref{fig:miro_model_config_low} and~\ref{fig:miro_model_config_high} show how end-to-end task latency changes under these configurations at different TPS.
We only show the results from \mirothinker for brevity; the trend was similar for \owl.
As expected, at TPS=0.1 (Figure~\ref{fig:miro_model_config_low}), allocating more GPUs to the \subllm's prefill (second and third plots) slightly improved overall latency over the balanced setup (first plot).
Once the \subllm's prefill is no longer the bottleneck, the \mainllm's decode becomes the next dominant cost.
Consequently, taking the GPUs from the \mainllm's prefill (third plot) yields slightly better performance than taking them from its decode (second plot), because the prefill stage has more slack to give up at this arrival rate.

The trend changes at TPS=0.5 (Figure~\ref{fig:miro_model_config_high}).
As before, taking GPUs from the \mainllm's decode and giving them to the \subllm's prefill (second plot) improves overall latency.
However, taking GPUs from the \mainllm's prefill (third plot) now has an adverse effect: end-to-end latency increases significantly, because the \mainllm's prefill suddenly becomes the bottleneck at this higher arrival rate.
The results suggest that a configuration that is optimal at one arrival rate can be highly sub-optimal at another.

\paragraph{\textbf{Takeaway 7}}
The optimal GPU allocation for prefill-decode disaggregation depends on the task arrival rate---a configuration that is near-optimal at one TPS can be substantially worse at another. This suggests that dynamic reconfiguration of GPU allocation, adjusting to the current load, could be beneficial.

\subsection{Simulating Larger Models} 
\label{subsec:scalability}

\begin{figure}[t]
\includegraphics[width=\columnwidth]{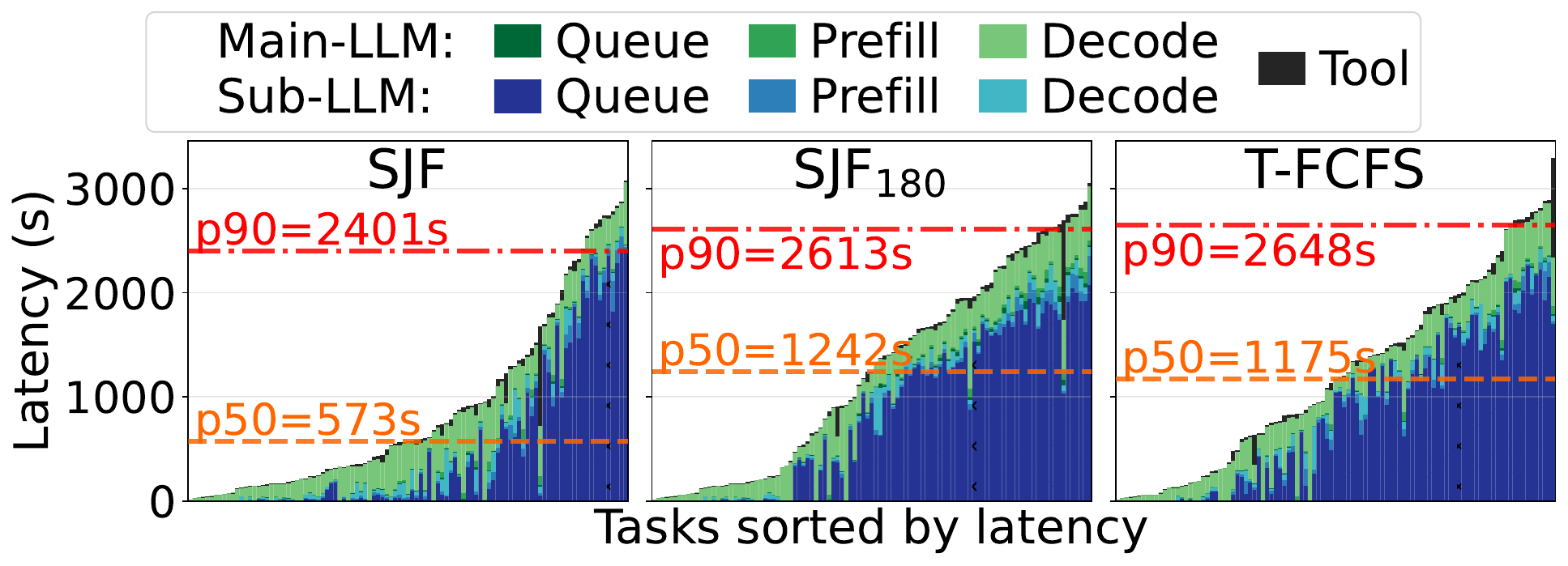}
\caption{Different scheduling policies with 2$\times$ larger model size (\mirothinker; otherwise same with Figure~\ref{fig:miro_scheduling_high}).
%\dhk{need TTFT TPOT for different HW config}
%\kwm{only leave SJF inf and SJF180. Add T-FCFS}
}
\label{fig:miro_scheduling_high_70B}
\end{figure}

\begin{figure}[t]
\includegraphics[width=\columnwidth]{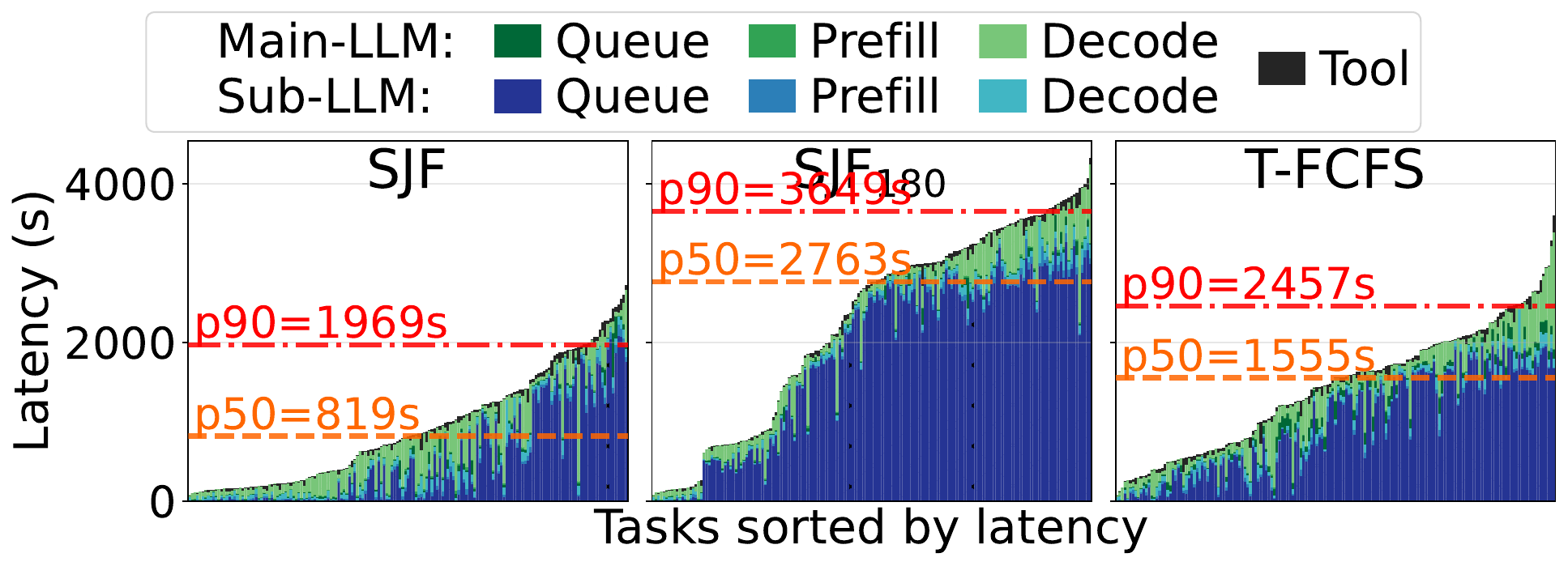}
\caption{Different scheduling policies with 2$\times$ larger model size (\owl; otherwise same with Figure~\ref{fig:owl_scheduling_high}).
}
\label{fig:owl_scheduling_high_70B}
\end{figure}

We reran the experiments from Figures~\ref{fig:miro_scheduling_high} and~\ref{fig:owl_scheduling_high} assuming a larger 70B model (Llama3-70B).
This is possible because \sys can simulate traces under an arbitrary model architecture, regardless of which model the trace was originally collected from---in effect, asking ``what if'' a similar trace had been generated by a different model.
Figures~\ref{fig:miro_scheduling_high_70B} and~\ref{fig:owl_scheduling_high_70B} show that SJF with a timeout ($\text{SJF}_{180}$) gives up much of plain SJF's latency advantage. In Figure~\ref{fig:owl_scheduling_high_70B}, it becomes substantially worse than both plain SJF and T-FCFS.
As discussed in Section~\ref{subsec:scheduling}, promoting a long-waiting query (which is typically prefill-heavy) evicts active KV cache entries belonging to other queries.
This adverse effect becomes more severe on larger models because GPU memory is relatively scarcer.
%, so each eviction discards a larger fraction of the active KV cache.
%
The effect is more pronounced in \owl, since \mirothinker's \subllm already gains little from the KV cache reuse.

\paragraph{\textbf{Takeaway 8}}
Scheduling behavior also changes with model size. In particular, the relative ranking of scheduling policies is not preserved across model sizes---on larger models, timeout-based promotion for fairness becomes less effective, since KV-cache eviction costs scale with model size.

% \subsection{KV Cache Eviction vs Recomputation} 
% \label{subsec:eviction}
% When GPU memory is exhausted, the serving system must either evict cached KV blocks to host memory (swap) or discard them entirely and recompute on demand.
% In agentic workloads, eviction is particularly costly because evicted blocks belong to long, multi-turn contexts whose recomputation cost scales linearly with accumulated prefill length.

% \subsection{Chunked Prefill} 
% \label{subsec:pdvschunked}

\subsection{Simulator Accuracy and Cost}
\label{subsec:correctness}

\sys predicts execution time with the operator-level model introduced by Vidur~\cite{agrawal_2024_vidur}, fitted to kernel measurements taken on the
target hardware. 
We validate it against a real Dynamo~1.1.0 + vLLM~0.19.0 prefill/decode-disaggregated deployment on four A100 (80GB) GPUs, replaying the same MiroThinker trace (2082 requests, 103 tasks).
All four setups (\texttt{S1}--\texttt{S4}) replay the same two-model workload. 
\texttt{S1} uses two GPUs for prefill (TP-2) and two GPUs for decode (TP-2), with the \mainllm and \subllm sharing the same prefill/decode GPUs (four GPUs in total). \texttt{S2} uses one GPU each for prefill and decode. Again, \mainllm and \subllm share the prefill/decode GPU (two GPUs in total). 
\texttt{S3} uses the same GPU setups with \texttt{S1} but uses Qwen2.5-32B instead of Llama-8B. 
\texttt{S4} gives one GPU each for \mainllm prefill, \mainllm decode, \subllm prefill, and \subllm decode  (four GPUs in total).
 Trace token IDs are clipped into the model's $128$K-entry vocabulary so the trace can be replayed through the real tokenizer, which preserves the prefix-sharing structure and cache hit rate.
We profiled every kernel the model needs under the same vLLM build the deployment runs, and release the profiles and profiling scripts so these numbers can be reproduced and regenerated for other hardware. 
%
%Our accuracy test compares the latency at task granularity.
%
%The workload is closed-loop, so each completion releases the task's next turn.
%
%The real and simulated runs drift into different arrival patterns, and per-request comparison would measure the two sides under different queue states.

The real and simulated runs experience different arrival patterns because task arrival order is randomized, causing each task to experience different queuing delays. Therefore, rather than comparing per-task completion latency directly, we compare the overall latency distributions using P50 and P90 per-task completion latency, geometric mean latency, and total time to complete all tasks, under simulated and real runs.
%However, our per-task latency comparison shows that each task’s completion time is closely predicted by the simulator.
%
Table~\ref{tab:correctness} shows that the median and geometric-mean task latencies of the simulated and real runs agree within $8\%$, and the total completion time differs by less than $4\%$. P90 latency exhibits somewhat larger variation, reaching $12.8\%$ in one case, but remains within $5\%$ for all other cases.
%Repeating \texttt{S1} at lower arrival rates leaves the error
%essentially unchanged. At TPS~$0.1$ the deployment is queue-dominated, while both $0.04$ and $0.01$ are decode dominated similar to Figure~\ref{fig:miro_tps}.

\begin{table}[t]\centering\small
\caption{Real and simulated task execution time (sec) and the simulation error, MiroThinker 103 task traces on up to 4 A100.}
\resizebox{\columnwidth}{!}{
\label{tab:correctness}
\begin{tabular}{llcccc}
\toprule
 & TPS & P50 & P90 & GeoMean & Total time \\
\midrule
\texttt{S1}    & 0.1  & 381/354 (-7.0\%) & 744/738 (-0.8\%)   & 306/282 (-7.6\%) & 2035/2040 (+0.3\%) \\
\texttt{S1}    & 0.04 & 204/196 (-3.7\%) & 437/381 (-12.8\%)  & 161/150 (-7.0\%) & 2745/2649 (-3.5\%) \\
\texttt{S1}    & 0.01 & 157/151 (-3.8\%) & 323/322 (-0.2\%)   & 125/118 (-6.3\%) & 9547/9207 (-3.6\%) \\
\texttt{S2}    & 0.1  & 771/745 (-3.4\%) & 1629/1566 (-3.9\%) & 572/534 (-6.6\%) & 2783/2769 (-0.5\%) \\
\texttt{S3}    & 0.1  & 2751/2669 (-3.0\%) & 5373/5187 (-3.5\%) & 1798/1738 (-3.3\%) & 6484/6308 (-2.7\%) \\
\texttt{S4}    & 0.1  & 315/314 (-0.4\%) & 731/764 (+4.6\%)   & 229/241 (+5.4\%) & 1847/1881 (+1.9\%) \\

\bottomrule
\end{tabular}
}
\end{table}

Simulation using \sys requires only a moderate CPU and can be done with low overheads. 
Getting the real-run numbers in Table~\ref{tab:correctness} requires running 2--4 A100s for $31$--$159$ \emph{minutes}, while getting the simulated numbers using \sys takes only $173$--$277$ \emph{seconds} on AMD EPYC 7763 ($9$--$35\times$ faster, with much less compute resources). 
%
%Simulations are independent, so one 
%machine runs them concurrently and the whole table costs about as much as its slowest entry.

%\input{Chapters/5_related}
\section{Conclusion}
\label{sec:Conclusion}

We presented \dataset and \sys, a trace dataset and trace-driven simulator that together enable the study of complex agentic AI  handling a heterogeneous mix of general tasks.
Our initial characterization surfaced several findings, demonstrating the value of the new dataset and simulator.
Our \dataset and \sys are open-sourced and hope they serve as a foundation for future work on agentic AI systems.

\section*{Acknowledgements}

This work was partly supported by the U.S. National Science Foundation under award No. CNS-2349610 and CCF-2529883. Any opinions, findings, and conclusions or recommendations expressed in this material are those of the author(s) and do not necessarily reflect the views of the National Science Foundation. 
The authors used Claude for code, text, and figures, but all outputs were reviewed carefully by the authors.

%-------------------------------------------------------------------------------
\bibliographystyle{IEEEtranS}
\bibliography{reference}

% LaTeX template for Artifact Evaluation V20240722
%
% Prepared by Grigori Fursin with contributions from Bruce Childers,
%   Michael Heroux, Michela Taufer and other colleagues.
%
% See examples of this Artifact Appendix in
%  * ASPLOS'24 "PyTorch 2: Faster Machine Learning Through Dynamic Python Bytecode Transformation and Graph Compilation": 
%      https://dl.acm.org/doi/10.1145/3620665.3640366
%  * SC'17 paper: https://dl.acm.org/citation.cfm?id=3126948
%  * CGO'17 paper: https://www.cl.cam.ac.uk/~sa614/papers/Software-Prefetching-CGO2017.pdf
%  * ACM ReQuEST-ASPLOS'18 paper: https://dl.acm.org/citation.cfm?doid=3229762.3229763
%
% (C)opyright 2014-2024 cTuning.org
%
% CC BY 4.0 license
%

\newpage
\appendix
\section{Artifact Appendix}

%%%%%%%%%%%%%%%%%%%%%%%%%%%%%%%%%%%%%%%%%%%%%%%%%%%%%%%%%%%%%%%%%%%%%
\subsection{Abstract}
This artifact studies how agentic AI workloads behave on an LLM serving cluster.
GAIATrace is a request-level execution traces recorded from two agent systems, OWL~\cite{hu_2025_owl} and MiroThinker~\cite{miromind_2026_mirothinker}, running the GAIA~\cite{mialon_2023_gaia} benchmark, plus the scripts that derive the paper's characterization results from them.
Vidur-Agent replays those traces against a simulated serving cluster, extending the Vidur simulator with heterogeneous replica groups, prefill–decode disaggregation, dependency-released multi-turn sessions.

\subsection{Artifact check-list (meta-information)}

%{\em Obligatory. Use just a few informal keywords in all fields applicable to your artifacts
%and remove the rest. This information is needed to find appropriate reviewers and gradually 
%unify artifact meta information in Digital Libraries.}

{\small
\begin{itemize}
  % \item {\bf Algorithm: }
  % \item {\bf Program: }
  % \item {\bf Compilation: }
  % \item {\bf Transformations: }
  % \item {\bf Binary: }
  % \item {\bf Model: }
  \item {\bf Data set: } GAIA 2023 validation set
  \item {\bf Run-time environment: } Ubuntu 24.04, Python 3.10, uv
  \item {\bf Hardware: } x86-64
  % \item {\bf Run-time state: } 
  % \item {\bf Execution: }
  % \item {\bf Metrics: }
  \item {\bf Output: } Figures, text tables, metrics
  \item {\bf Experiments: } Trace characterization and simulation
  \item {\bf How much disk space required (approximately)?: } 10GB
  \item {\bf How much time is needed to prepare workflow (approximately)?: } 20 minutes
  \item {\bf How much time is needed to complete experiments (approximately)?: } 20 minutes
  \item {\bf Publicly available?: } Yes
  \item {\bf Code licenses (if publicly available)?: } Apache-2.0 (GAIATrace), MIT(Vidur-Agent)
  \item {\bf Data licenses (if publicly available)?: } CC-BY-4.0 (traces)
  % \item {\bf Workflow automation framework used?: }
  \item {\bf Archived (provide DOI)?: } Yes
\end{itemize}
}

%%%%%%%%%%%%%%%%%%%%%%%%%%%%%%%%%%%%%%%%%%%%%%%%%%%%%%%%%%%%%%%%%%%%%
\subsection{Description}

\subsubsection{How to access}

Both GAIATrace and MiroThinker are published in Zenodo as sibling directories: \\
(https://doi.org/10.5281/zenodo.22133686)

\subsubsection{Hardware dependencies}
Any x86-64 machine; no GPU, since operator execution times come from the shipped profiling data.
The sweeps run concurrently and predictor fitting uses all cores.

\subsubsection{Software dependencies}
We used Ubuntu 24.04, Python 3.10, detailed instruction in 
Appendix~\ref{subsec:install}.

\subsubsection{Datasets}
This artifact ship are the traces recorded from the two agent systems represented in token\_id. GAIA benchmark is not shipped.
To make validation set not be republished in crawlable form, text verbatim files are left out. What ships is still enough to reproduce every result.
% The recorded traces are shipped, not GAIA itself.
% per-run CSVs at each tree's \texttt{traces/session\_traces/}, merged by \texttt{merge.py} into \texttt{owl\_random.csv} (5,406
% requests, 165 sessions) and \texttt{miro\_random.csv} (2,082 requests, 103 sessions).

% \subsubsection{Models}

%%%%%%%%%%%%%%%%%%%%%%%%%%%%%%%%%%%%%%%%%%%%%%%%%%%%%%%%%%%%%%%%%%%%%
\subsection{Installation}
\label{subsec:install}
Install from Zenodo: \\
(https://doi.org/10.5281/zenodo.22133686)

\begin{verbatim}
pip install -r GAIATrace/owl/requirements.txt
cd Vidur-Agent
uv venv && source .venv/bin/activate && uv sync
\end{verbatim}

\subsection{Experiment workflow}

\begin{verbatim}
# Stage 1 (GAIATrace): Trace gen + Characterization
cd GAIATrace
./GAIATrace/owl/run_all.sh       
./GAIATrace/mirothinker/run_all.sh

# Stage 2 (Vidur-Agent): Trace Simulation 
cd ../Vidur-Agent
# Fit predictors once
bash warmup_predictors.sh        

# Simulator runners
bash run_miro.sh && bash run_owl.sh
bash run_miro70B.sh && bash run_owl70B.sh
bash analysis/run_gantt_compare.sh
bash analysis/run_plot_ttft_tpot.sh
\end{verbatim}

Predictor fitting (\texttt{warmup\_predictors.sh}) saturates every core, so it trains one configuration at a time, covering every model and parallelism the sweeps use. 
The sweep scripts launch their runs concurrently by default; on machines with few cores or small DRAM, sequentialize them to fit the available capacity.

%%%%%%%%%%%%%%%%%%%%%%%%%%%%%%%%%%%%%%%%%%%%%%%%%%%%%%%%%%%%%%%%%%%%%
\subsection{Evaluation and expected results}

GAIATrace writes to each tree's \texttt{analysis/outputs/}, generating
Figure\ref{fig:miroflow_scatter}--Figure\ref{fig:other_samples}, Table\ref{tab:trace_stats} and Table\ref{tab:tool_stats}. 
Vidur-Agent writes per-run metric CSVs. 
Running \texttt{analysis/*.sh} gives \texttt{analysis/gantt\_compare/*.pdf} and
\texttt{analysis/ttft\_tpot/*.pdf} corresponds to Figure~\ref{fig:prefix_caching_miro}--Figure~\ref{fig:owl_scheduling_high_70B}. Simulation is deterministic given a trace, cluster configuration and seed, so the outputs reproduce exactly.

%%%%%%%%%%%%%%%%%%%%%%%%%%%%%%%%%%%%%%%%%%%%%%%%%%%%%%%%%%%%%%%%%%%%%
\subsection{Experiment customization}
\begin{itemize}
    \item \textbf{Workload} --- any CSV with \texttt{num\_prefill\_tokens} and \texttt{num\_decode\_tokens} are accepted. Additional columns like \texttt{session\_id}, \texttt{dep},
    \texttt{inter\_request\_latency} and \texttt{token\_ids} add multi-turn structure, dependencies, tool time and token-exact prefix caching.
  \item \textbf{HW Configuration: }
    \texttt{replica\_groups\_configs/} files declare replica groups (role, count, model, parallelism, device) and, per model, which replicas prefill and which decode. Need additional warmup when testing uncached hardware configurations.
  % \item \textbf{Task arrival rate} 
  \item \textbf{Scheduling policy:} Request scheduling policies like session priority and starvation prevention for per-replica admission.
  
\end{itemize}

%%%%%%%%%%%%%%%%%%%%%%%%%%%%%%%%%%%%%%%%%%%%%%%%%%%%%%%%%%%%%%%%%%%%%
\subsection{Notes}
The READMEs at the repository root and in each subdirectory document the pipeline, trace format, cluster configuration schema.

\end{document}